\documentclass[preprint,12pt]{elsarticle}

\usepackage{amssymb}
\usepackage{graphicx}
\usepackage{amssymb}
\usepackage{amsmath}
\usepackage{bm}
\usepackage[caption=false,font=normalsize]{subfig}
\usepackage[colorlinks,citecolor=blue,linkcolor=green]{hyperref}

\begin{document}

\begin{frontmatter}



\title{Continuous Trajectory Planning Based on\\ Learning Optimization in High Dimensional Input Space \\for Serial Manipulators}


\author{Shiyu Zhang, Shuling Dai}

\address{Xueyuan Road No.37, Haidian District, Beijing, China}

\begin{abstract}
To continuously generate trajectories for serial manipulators with high dimensional degrees of freedom (DOF) in the dynamic environment, a real-time optimal trajectory generation method based on machine learning aiming at high dimensional inputs is presented in this paper. First, a learning optimization (LO) framework is established, and implementations with different sub-methods are discussed. Additionally, multiple criteria are defined to evaluate the performance of LO models. Furthermore, aiming at high dimensional inputs, a database generation method based on input space dimension-reducing mapping is proposed. At last, this method is validated on motion planning for haptic feedback manipulators (HFM) in virtual reality systems. Results show that the input space dimension-reducing method can significantly elevate the efficiency and quality of database generation and consequently improve the performance of the LO. Moreover, using this LO method, real-time trajectory generation with high dimensional inputs can be achieved, which lays a foundation for continuous trajectory planning for high-DOF-robots in complex environments.
\end{abstract}


\end{frontmatter}


\section{Introduction}
When working in the a dynamic environment, for example, playing ball games \cite{ref:ren15}, catching flying objects \cite{ref:kim14,ref:salehian16} and performing haptic feedback, robots are required to quickly react to the continuous changes of environment and objectives, for which continuous trajectory planning is needed. In frequently changing environments, trajectories need to be generated in a very short period, which requires to reduce the calculation time for trajectory planning.

Trajectory planning methods based on non-linear optimization are always adopted to improve the motion efficiency by optimizing criteria such as motion time, energy and power consumption. However, the objective functions and constraints are always very complex in practical problems, leading to non-linear and non-convex optimization problems, which are likely to stuck into local minimum and very time-consuming to solve. Solving global minimum of non-linear optimization problems in real-time is still a challenge.

Learning from the former data is a promising way to reduce on-line calculation time. First, plenty of non-linear optimization problems are solved off-line to establish a database of optimal trajectories. Then, new optimal trajectories are generated on-line by regression and prediction. Two key factors are in the learning framework, the database and the learning algorithm.

The database is the foundation of learning, which is the precondition of learning algorithms and mainly determines the learning performance. The inappropriate database may lead to poor performance of learning, then the low working efficiency of robots, even the failure to finish the task. Thus, database generation is of crucial importance. The traditional way resort to randomly or evenly choosing sample variables in the input space and calculating the global minimum by brute force methods to obtain the samples. However, in the cases of complex environments or robots with high DOFs, the required sample amount exponentially increases, which greatly increases the calculation cost. Moreover, many invalid or unrepresentative samples may be generated in this way, which may limit the learning performance.

On-line learning is another important part of learning, which is the key to make full use of the database. A range of problems need to be tackled in on-line learning. First, traditional learning algorithms can only handle cases with single output. Regression model establishment and feature selection for cases with high dimension inputs and multiple outputs are the major issues. In addition, optimization problems in different applications have different requirement of performance, which requires to select appropriate regression algorithms in specific applications. Furthermore, ordinary learning algorithms have no guarantee that the output parameters satisfy the constraints. Thus, for constrained optimization problems, further refinements for the parameters output by common learning algorithms are needed.

Considering continuous trajectory planning problems for serial manipulators, a learning optimization (LO) framework is constructed in this paper. In particular, aiming at high dimensional inputs, an input space dimension-reducing mapping method is presented, to reduce the cost and improve the database quality in database generation. Moreover, multivariable-multiple-regression model is constructed to deal with learning problems with multiple inputs and multiple outputs. In addition, multiple performance evaluation indices are defined to measure the feasibility, real-time performance and accuracy of the LO model. At last, the above method is applied to the specific case, motion planning of haptic feedback manipulators, to verify the theory.

\section{Related Work}
The continuous trajectory planning problem for robots, which is to re-plan the trajectory for robots in a high frequency according to the changes of obstacles and the objective in dynamic environments, is widely used in practice, such as playing ball games \cite{ref:ren15} and catching flying objects \cite{ref:kim14,ref:salehian16}. It plays an important role as the foundation for human-robot interaction, like collaborative manufacturing \cite{ref:mainprice16,ref:iqbal16} and playing juggling \cite{ref:kober12}. The continuous trajectory planning problem can be split into a series of point-to-point trajectory planning problems, which is high-demanding in real-time performance.

The traditional point-to-point trajectory planning is started from interpolation-based methods, such as polynomial interpolation \cite{ref:kim97,ref:bazaz97} and B-spline interpolation \cite{ref:lampariello13,ref:liu15}. In general, pure interpolation-based methods are able to accomplish required tasks, but difficult to achieve the optimal performance in specific aspects. To obtain optimal trajectories, non-linear optimization problems are constructed with optimal objectives based on time, energy and power consumption and constraints such as mechanical structure, time and obstacle avoidance \cite{ref:schulman14}. Von Stryk et al. \cite{ref:von94} investigated the non-linear optimization with three separated criteria of minimum time, minimum energy and minimum power consumption and solved it by a numerical method of combining a direct collocation and an indirect multiple shooting method. Chettibi et al. \cite{ref:chettibi04} presented the optimal planning problem tried to find the compromise between time, energy and power consumption and solved it by the Sequential Quadratic Programming (SQP) method. However, all aforementioned optimization-based methods are not real-time due to the complex computation of the non-linear optimization.

The non-linear optimization is likely to get stuck into local minimum and generally solved with multiple initial guesses to obtain the global minimum, which is significant costly and hard to operate in real-time. Therefore, quickly seeking for the global minimum is still challenging. A promising idea is learning from former data to reduce the on-line calculation time \cite{ref:ude10,ref:cassioli12,ref:pan14}. Lampariello et at. \cite{ref:lampariello11} and Werner et al. \cite{ref:werner15} solved the non-linear optimization by optimal database generation off-line and regression prediction on-line and applied it in catching flying objects and bipedal walking. However, the methods are limited in specific applications and cases with low dimensional inputs. Jetchev et al. \cite{ref:jetchev13} proposed a trajectory prediction method, employing the previous data to speed up generating new trajectories. In addition, high dimensional inputs are considered and a high dimensional situation descriptor is defined. Hauser \cite{ref:hk17} extended this method by considering more general context of nonlinear optimization problems and presented a general learning global optima (LGO) framework.

The database for learning can be obtained by either recording the former data, or artificially generating. Lampariello et al. \cite{ref:lampariello11} evenly chose variables in motion range to generate samples and drew a comparison of databases with different sizes. Hauser \cite{ref:hk17} uniformly sampled an axis-aligned range of variables to generate a database. Besides, a lifelong learning mode was presented to continuously generate examples with a separate background thread. In addition, the sensitivity of the required database size to the input dimension and the requirement for the database to guarantee the quality of solutions were discussed. However, for the cases with high dimensional inputs, the above database generation methods of randomly or evenly choosing sample variables in the motion range are time-consuming. Thus, finding a way to elevate the efficiency of database generation is a key issue to be solved.

This paper extends the LGO framework and constructs a learning optimization (LO) model, which can be adjusted by adopting different databases, regression methods and refinement methods to adapt different applications, together with the performance evaluation indices. Moreover, aiming at the low efficiency of database generation with traditional methods for high dimensional input space, a database generation method based on dimension-reducing mapping is presented. Finally, this method is validated in continuous trajectory planning for haptic feedback manipulators in virtual reality systems.

\section{Continuous Trajectory Planning Based on Learning Optimization}
\subsection{Continuous Trajectory Planning}    \label{sec:cplanning}
Considering trajectory planning problems in joint space, the trajectory parameters are represented as $\boldsymbol{C} \in \mathbb{R}^{N_C}$, and then the position, velocity, and acceleration are represented as $\boldsymbol{q}(\boldsymbol{C},t)$, $\boldsymbol{\dot{q}}(\boldsymbol{C},t)$ and $\boldsymbol{\ddot{q}}(\boldsymbol{C},t)$, respectively. The input of the trajectory problem is represented as $\boldsymbol{X} \in \mathbb{R}^{N_X}$, including the initial configuration, the objective configuration, the initial velocity, the environmental variables etc. The trajectory parameters are determined by the input state. In essence, the trajectory planning problem is the projection from input variables $\boldsymbol{X}$ to trajectory parameters $\boldsymbol{C}$
\begin{equation}
\boldsymbol{X} \to \boldsymbol{C}=f(\boldsymbol{X}).
\end{equation}

\begin{figure}[h]
 \centering
 \includegraphics[width=0.5\columnwidth]{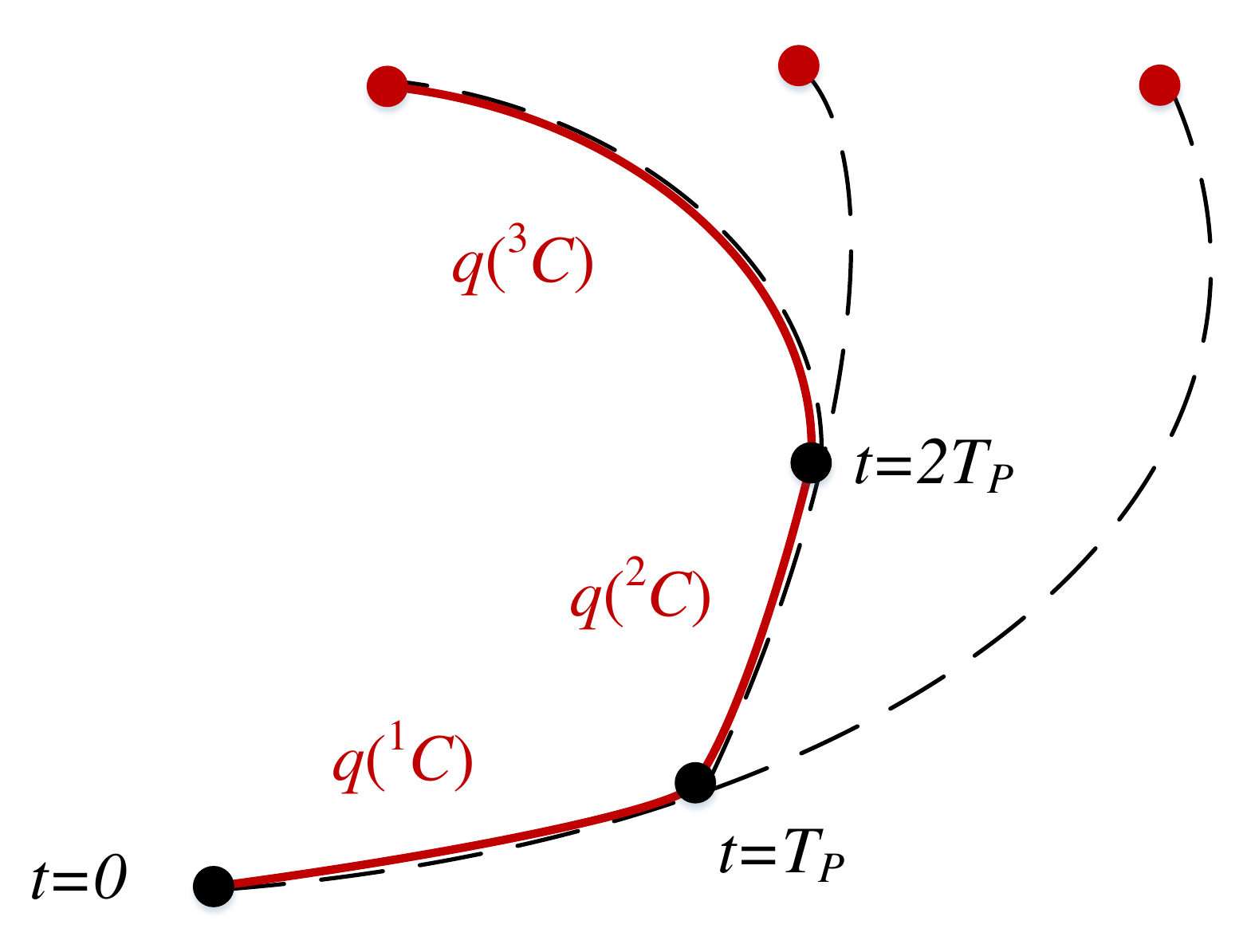}
 \caption{Continuous trajectory planning (black lines: trajectories generated in each period, red line: the actual trajectory of the whole process)}
 \label{fig:traj}
\end{figure}

When serial manipulators work in a dynamic environment, in which the objective configuration or the environmental obstacle positions change all the time, $\boldsymbol{X}$ changes frequently. It is required to re-plan the trajectory according to new $\boldsymbol{X}$s. In the $i$th detection period $T_p$, we detect the environment and update input to obtain $\boldsymbol{X^i}$, then re-plan the trajectory and obtain the according parameters $\boldsymbol{C^i}$, until the robot reaches the actual objective configuration, shown as Figure \ref{fig:traj}. The trajectory of the whole procedure is
\begin{equation}
\boldsymbol{q}(\boldsymbol{C^i},t), t \in [(i-1)T_p,iT_p), i \in [1,N_T],
\end{equation}
where $N_T$ is the total number of planning periods.

To guarantee the accuracy of trajectory planning, $T_p$ is always very small. Thus, real-time performance of trajectory planning is required, which means that the calculation time is shorter than $T_p$.

\subsection{Real-time Trajectory Planning Based on Learning Optimization}
\subsubsection{Learning Optimization Framework} \label{sec:lo-model}
To improve the working efficiency of robots, non-linear optimization based method is employed to generate the trajectory, in which the objective function $F(\boldsymbol{C})$ is defined to minimize the energy consumption and the motion time of the robot and the constraints are determined by considering the mechanism limitations, time limitations and other limitations related to the task. Optimal parameter $\boldsymbol{C}^\ast$ can be obtained by solving the optimization problem to minimize $F(\boldsymbol{C})$ while satisfy all the constraints. The optimization problem is specified by $\boldsymbol{X}$:
\begin{equation}
\begin{split}
\boldsymbol{C}^\ast=\min_{\boldsymbol{C}} F_X(\boldsymbol{C}) \\
s.t.\\
\boldsymbol{C} \in \mathbb{R}^{N_{C}},   \\
{h_X}^i(\boldsymbol{C})=0, (i=1,2,\ldots,N_h),      \\
{g_X}^j(\boldsymbol{C})\ge 0, (j=1,2,\ldots,N_g),      \\
\end{split}
\end{equation}
which is denoted as
\begin{equation}
\boldsymbol{C}^\ast=O(\boldsymbol{X}),   \label{equ:optim}
\end{equation}
where $h$ and $g$ represent the equation constraints and the inquation constraints, respectively.

\begin{figure}[h]
 \centering
 \includegraphics[width=0.7\columnwidth]{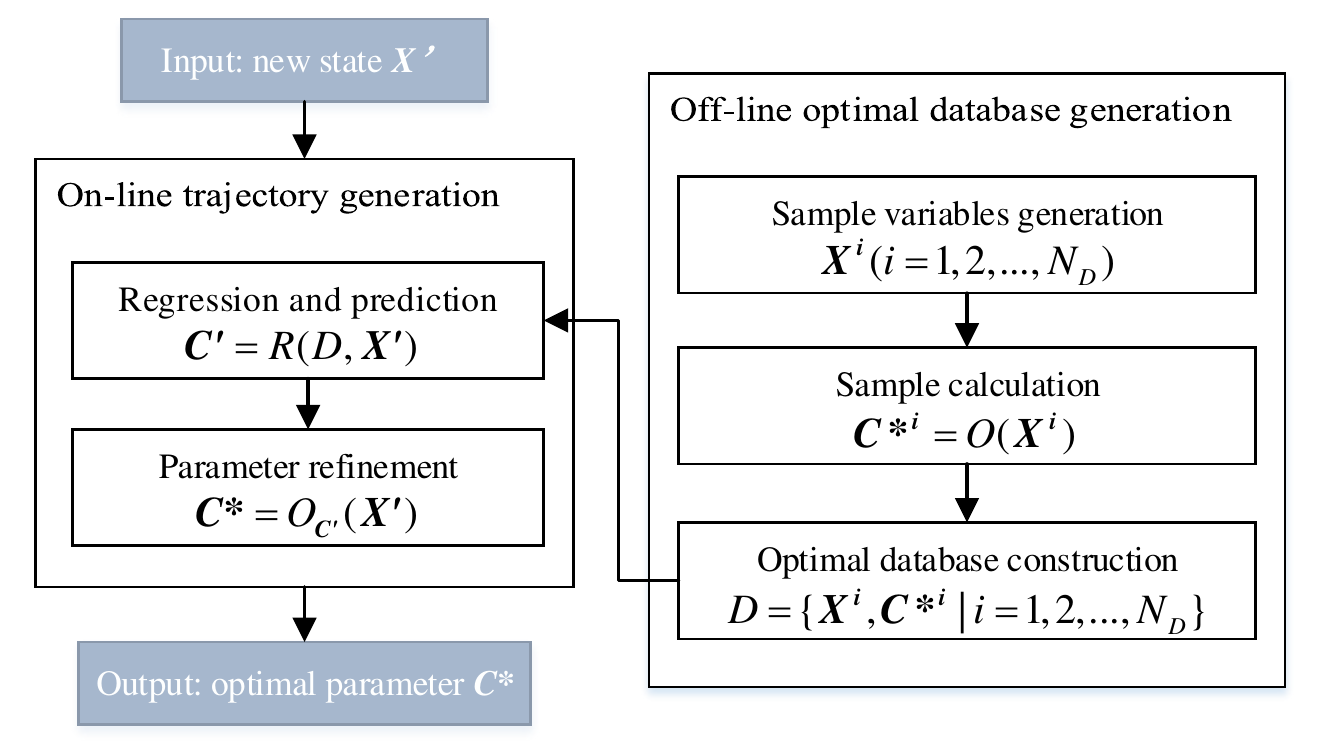}
 \caption{Real-time trajectory generation based on learning optimization}
 \label{fig:loflow}
\end{figure}

In general, the above non-linear optimization problem is non-convex, which is likely to get stuck into local minimum, making the global optimization time-consuming and hardly to be solved in real-time. In order to achieve real-time performance, we employ machine learning methods to solve the non-linear optimization problem by learning new solutions from the optimal trajectory database, which is generated off-line. The on-line procedure includes regression prediction and parameter refinement. The detailed learning optimization method is as following (shown as Figure \ref{fig:loflow}).

\begin{enumerate}
  \item Database generation

Select multiple input variables $\boldsymbol{X^i} (i=1,2,¡­,N_D)$ in the input space and solve the global minimum by the multi-restart method
\begin{equation}
{\boldsymbol{C}^\ast}^i=\boldsymbol{X^i}.
\end{equation}

Build the optimal database
\begin{equation}
D=\{\boldsymbol{X^i},{\boldsymbol{C}^\ast}^i| i=1,2,¡­,N_D \}.
\end{equation}

  \item Regression and prediction

Establish the regression model by exploiting the optimal database, which can be performed off-line or on-line. Specifically, if the regression model is related to the new input state $\boldsymbol{X'}$, such as $k$-NN, the regression model is established on-line. Otherwise, the regression model establishment is performed off-line, such as SVR and GPR.

According to the new input state $\boldsymbol{X'}$, the initial parameter can be predicted with the regression model
\begin{equation}
\boldsymbol{C'}=R(D,\boldsymbol{X'}).
\end{equation}

  \item Parameter refinement

The initial parameter $\boldsymbol{C'}$ is not guaranteed to satisfy the constraints. Thus, we take $\boldsymbol{C'}$ as the initial parameter to solve an optimization problem, which has the same constraints with the original optimization problem to refine the parameters
\begin{equation}
\boldsymbol{C}^\ast=O_{\boldsymbol{C'}}(\boldsymbol{X'}).
\end{equation}

\end{enumerate}

The above trajectory planning method based on optimization and machine learning is mainly determined by three elements, database $D$, regression and prediction method $R$ and parameter refinement method $O$. To explore the individual influences of the three elements, we define them as sub-methods and construct the learning optimization (LO) model
\begin{equation}
\mathcal{L}(D,R,O).
\end{equation}

With different sub-methods, $\mathcal{L}$ have different learning performance and consequently can be used in different applications. The LO model can be evaluated by several indices, which are detailed in Section \ref{sec:lo-indices}.

\subsubsection{Performance Evaluation Indices} \label{sec:lo-indices}
To evaluate feasibility, real-time performance and accuracy of different $\mathcal{L}$s specified by $(D,R,O)$, we define three performance evaluation indices for $\mathcal{L}$, which are determined by $(D,R,O)$ as well.

\begin{enumerate}
  \item Feasible success rate

According to \ref{sec:lo-model}, for a new input $\boldsymbol{X'}$, an initial parameter $\boldsymbol{C'}$ is obtained by regression and prediction, which needs a further refinement by solving a non-linear optimization problem. Assuming $N_r$ $\boldsymbol{X'}$s are input to $\mathcal{L}$, among which only $N_{r1}$ groups output feasible parameters in the refinement stage, the feasible success rate $R_s$ of $\mathcal{L}$ can be approximate as
\begin{equation}
R_s=N_{r1}/N_r.
\end{equation}

Feasible success rate is the most crucial index in LO, which indicates the feasibility of the LO model and reflects whether the task can be done successfully. In continuous trajectory planning, if we get an unfeasible solution in any planning period, such as a parameter does not satisfy the constraints of the non-linear optimization problem, the task cannot be fulfilled.

  \item Learning time

Learning time $T_L$ is the average time for $\mathcal{L}$ to perform an on-line learning, including regression prediction time $T_{PR}$ and refinement time $T_{RO}$:
\begin{equation}
T_L=T_{PR}+T_{RO}.
\end{equation}

It reflects weather trajectory planning can be performed in real-time. According to Section \ref{sec:cplanning}, in continuous trajectory planning applications, to achieve real-time performance, one on-line trajectory planning procedure is required to be finished in one prediction period $T_p$. Thus, $T_L$ is required to satisfy
\begin{equation}
T_L<T_p.
\end{equation}

  \item Cost increase rate

Assume $\boldsymbol{C_a}$ is the accurate optimal parameter, i.e., the solution of the global optimization problem, and $F_a$ is the corresponding objective function value. Similarly, $\boldsymbol{C_p}$ is the parameter derived through $\mathcal{L}$ and $F_p$ is the corresponding objective function value. Suppose we have $N_r$ groups of $(\boldsymbol{X},\boldsymbol{C_a},F_a)$, and run the LO to obtain $\boldsymbol{C_p}$ and $F_p$. Then the cost increase rate can be approximate as
\begin{equation}
e_F=\frac{1}{N_r} \sum_{i=1}^{N_r}|\frac{F_p^i-F_a^i}{F_a^i}|.
\end{equation}

It indicates the deviation of the performances from the optima, which does not influence the feasibility of the task, but reflects whether the task will be done in a high quality and efficiency.

\end{enumerate}

\section{Database Generation Method Based on Dimension-Reducing Mapping} \label{sec:database}
\subsection{Dimension-Reducing Mapping} \label{sec:database-map}
In the LO framework, ample samples are required to establish the database with high quality. Traditional method is randomly or evenly choosing sample variables in the changing range, that is, choosing some values in the changing range of each component of the input variable respectively and combining them to obtain sample vectors. Supposing $d$ values are chosen for each component, which is defined as variable density, then $d^{N_X}$ samples need to be calculated. When the dimension of the input variable $\boldsymbol{X}$ is high, i.e., $N_X$ is large, the required number of samples increases exponentially, which greatly increases the calculation cost. Furthermore, the sample variables chosen by this method are not always in the workspace of the specific application. Sample variables out of the workspace are less significant for learning. Moreover, some of them may be meaningless in Equation \ref{equ:optim}, which cannot converge to a feasible solution. Therefore, the high calculation cost and the existence of invalid samples lead to low efficiency and poor quality of database generation by using the traditional method, which significantly limits the performance of LO.

To improve the efficiency and quality of database generation for LO with high dimensional inputs, a database generation method based on input space dimension-reducing mapping is presented in this paper. The original input variables are mapping into a new variable space, which is help to select sample variables.

There are two main functions of mapping. One is to map the original input variables into a new variable space in which we can easily determine whether an input variable is in the workspace. The other is to reduce the dimension of input variables, which can help to improve the flexibility of sample selection and obtain better sample distribution.

The input variable $\boldsymbol{X}$ is mapped into $\boldsymbol{X_L}$
\begin{equation}
\begin{split}
\boldsymbol{X_L}=\boldsymbol{\mathcal{F}}(\boldsymbol{X}),    \\
\boldsymbol{\mathcal{F}}: \mathbb{R}^{N_X} \to \mathbb{R}^{N_{X_L}},
\end{split}
\end{equation}
where $\boldsymbol{\mathcal{F}}$ denotes the mapping, and $N_{X_L}$ represents the dimension of the new variable space.
Two types of mapping are illustrated in this paper, including merging mapping $\boldsymbol{\mathcal{F}_M}$ and transforming mapping $\boldsymbol{\mathcal{F}_T}$:
\begin{equation}
\boldsymbol{\mathcal{F}}=\{\boldsymbol{\mathcal{F}_M},\boldsymbol{\mathcal{F}_T}\}.
\end{equation}

\subsubsection{Merging Mapping} \label{sec:database-mapm}
A merging mapping $\boldsymbol{\mathcal{F}_M}$ is to map a high dimensional dependent component vector $\boldsymbol{x_{id}} \in \mathbb{R}^{N_{id}}$ into a low dimensional independent component vector $\boldsymbol{x_d} \in \mathbb{R}^{N_d}$
\begin{equation}
\begin{split}
\boldsymbol{x_{id}}=\boldsymbol{\mathcal{F}_M}(\boldsymbol{x_d}), \\
\boldsymbol{\mathcal{F}_M}: V_d \subset \mathbb{R}^{N_{id}} \to V_{id} \subset \mathbb{R}^{N_d}, \\
N_d>N_{id}.
\end{split}
\end{equation}

Merging mappings reduce the dimension of original input variables by combining the independent redundant elements and turned them into a dependent component vector, which can be used for solving non-linear optimization problem directly.

The variables used to calculate samples actually are $\boldsymbol{x_{id}}$s, instead of $\boldsymbol{x_d}$s. Thus, the sample variables can be randomly or evenly selected in the changing range of $\boldsymbol{x_{id}}$, which is detailed as follow:
\begin{enumerate}
  \item Regard the changing range of $\boldsymbol{x_d}$ as the domain of $\boldsymbol{\mathcal{F}_M}$, and solve for the range $V_{id}$, which is the changing range of $\boldsymbol{x_{id}}$;
  \item Randomly or evenly select variable vectors in $V_{id}$ with variable density $d$, and a set of sample variables $S=\{\boldsymbol{x_{id}^i} \in V_{id}| i=1,2,\ldots,d^{N_d}\}$ is obtained.
\end{enumerate}

\begin{figure}[h]
 \centering
 \includegraphics[width=0.5\columnwidth]{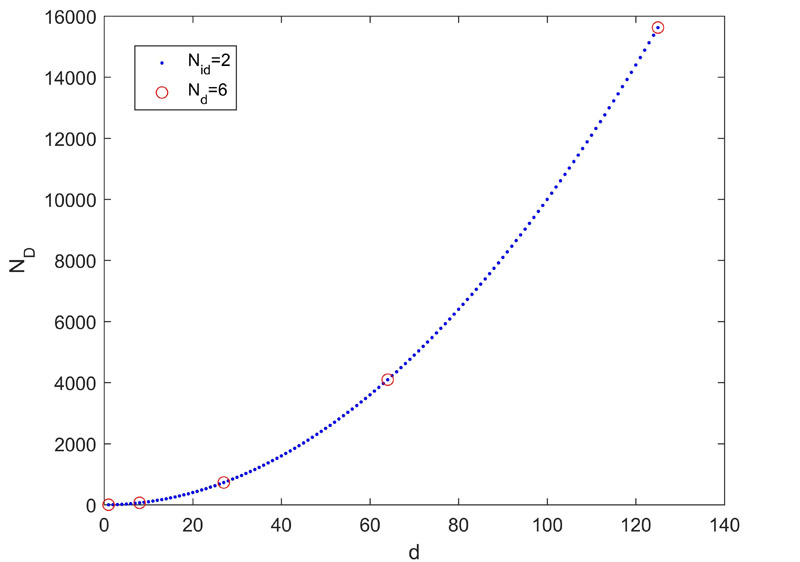}
 \caption{Probable sizes of the sample variable sets before and after dimension-reducing mapping}
 \label{fig:mapmsize}
\end{figure}

By above method, a sample variable set $S_1$ is obtained, which includes $N_D=d^{N_{id}}$ groups of sample variables. On the other side, if we adopt the traditional method by randomly or evenly selecting variable combinations of $\boldsymbol{x_d}$ in $V_d$ with variable density $d'=d^{N_{id}/N_d}$ and then mapping them into the space $\mathbb{R}^{N_{id}}$, another sample variable set $S_2$ is obtained, which also include $N_D=d^{N_{id}}$ groups of sample variables. However, as $N_{id}/N_d<1$ and the variable density is a positive integer, the probable sizes of $S_2$ is far less than $S_1$. For example, if $N_d=6$ and $N_{id}=2$, the probable sizes of the sample variable sets before and after dimension-reducing mapping are shown in Figure \ref{fig:mapmsize}.

Moreover, $S_1$ and $S_2$ has the same size, but different distribution of sample variables in $\mathbb{R}^{N_{id}}$. The sample variables in $S_1$ are always evenly distributed in $\mathbb{R}^{N_{id}}$ while in $S_2$ are trend to gather, which is less representative. For example, consider $d=27$ and $d'=3$, linear merging mapping $\boldsymbol{\mathcal{F}_{M1}}$ and non-linear merging mapping $\boldsymbol{\mathcal{F}_{M2}}$ are selected:
\begin{equation}
\boldsymbol{x_{id}}=\boldsymbol{\mathcal{F}_{M1}}(\boldsymbol{x_d})=
\left( \begin{array}{cccccc}
1 & 2 & 3 & 0 & 0 & 0 \\
0 & 0 & 0 & 0.5 & 3 & 10
\end{array} \right)
\boldsymbol{x_d},
\end{equation}
and
\begin{equation}
\begin{split}
\boldsymbol{x_d} \to \boldsymbol{x_{id}}=\boldsymbol{\mathcal{F}_{M2}}(\boldsymbol{x_d}): \\
x_{id}(1)={x_d}^3(1)+{x_d}^3(2)+{x_d}^3(3),  \\
x_{id}(2)={x_d}^3(4)+{x_d}^3(5)+{x_d}^3(6).
\end{split}
\end{equation}

\begin{figure}[h]
\centering
\subfloat[Linear mapping, uniform selection]{\includegraphics[width=0.4\textwidth]{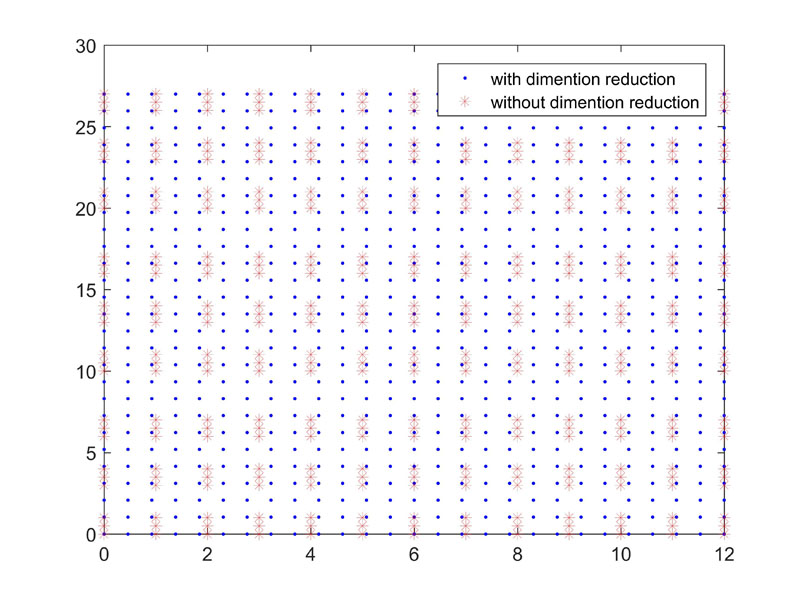}}
\hspace{2pt}
\subfloat[Non-linear mapping, uniform selection]{\includegraphics[width=0.4\textwidth]{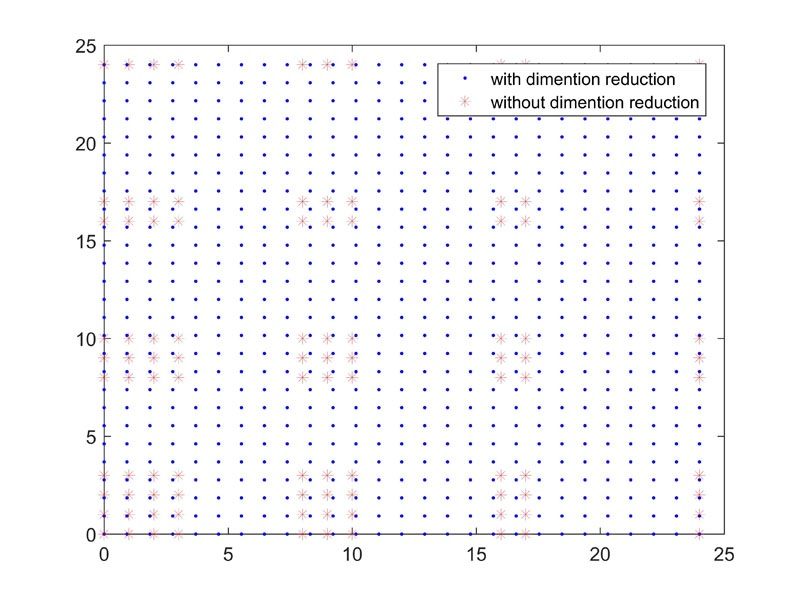}}
\hspace{2pt}
\subfloat[Linear mapping, random selection]{\includegraphics[width=0.4\textwidth]{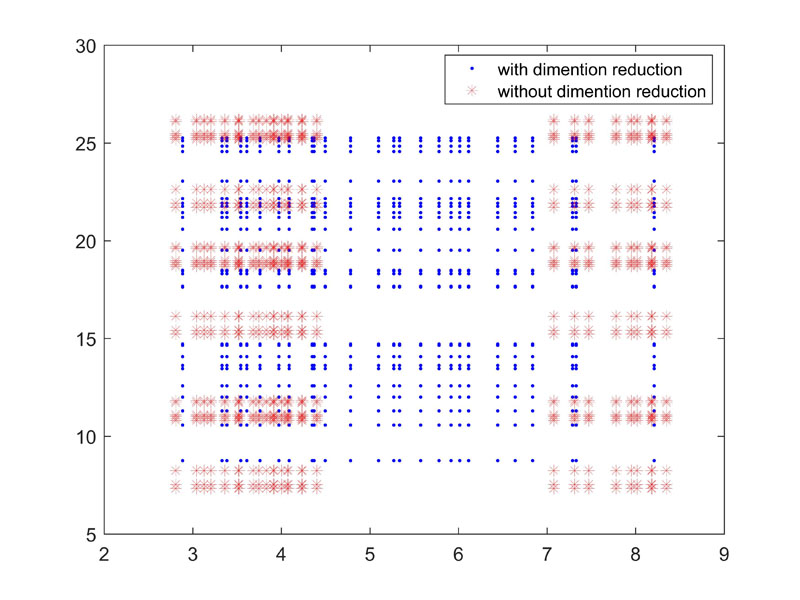}}
\hspace{2pt}
\subfloat[Non-linear mapping, random selection]{\includegraphics[width=0.4\textwidth]{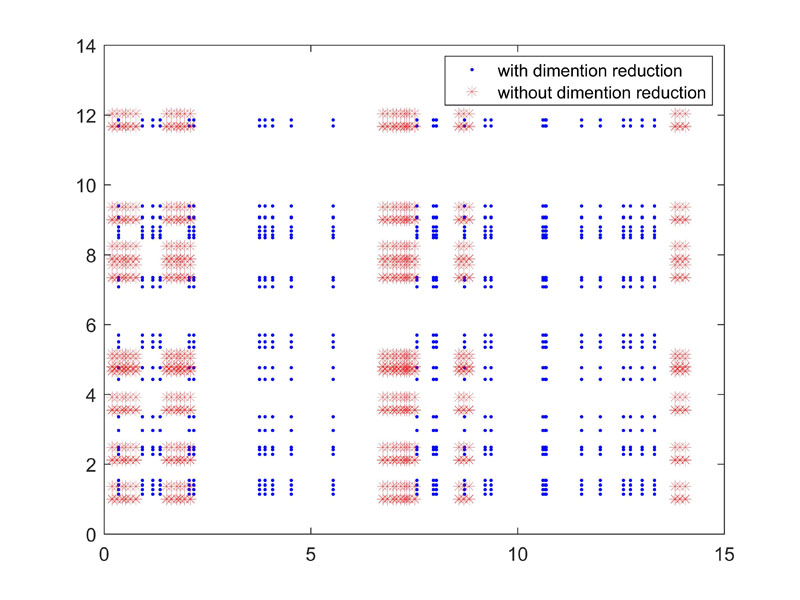}}
\caption{Sample variables distribution before and after dimension-reducing mapping}
\label{fig:mapmdist}
\end{figure}

The comparison of distributions of $\boldsymbol{x_{id}}$ obtained by randomly and uniformly selecting sample variables before and after dimension-reducing mapping are shown in Figure \ref{fig:mapmdist}.

\subsubsection{Transforming Mapping}    \label{sec:database-mapt}
A transforming mapping $\boldsymbol{\mathcal{F}_T}$ is to map a independent component vector of $\boldsymbol{X}$ into another variable space:
\begin{equation}
\begin{split}
\boldsymbol{x_{d2}}=\boldsymbol{\mathcal{F}_T}(\boldsymbol{x_{d1}}),   \\
\boldsymbol{\mathcal{F}_T}: V_{d1} \subset \mathbb{R}^{N_{d1}} \to V_{d2} \subset \mathbb{R}^{N_{d2}},   \\
N_{d1} \ge N_{d2}.	
\end{split}
\end{equation}

The dimension of the new variable space could be lower, or stay the same with the original component vector. The main aim of transforming mapping is to map the original input variables into the variable space in which we can determine whether an input variable is in the workspace.

Contrast to merging mapping, the variables used to calculate samples are the variables before transforming mapping $\boldsymbol{x_{d1}}$s. While $\boldsymbol{x_{d2}}$s are only used to determine whether the variable is in the workspace or not. Therefore, the variables are first selected in $V_{d1}$ and then mapped to $V_{d2}$ to perform workspace verification.

The specific procedure for generating sample variables is as follow.

\begin{enumerate}
  \item Randomly or evenly select multiple vectors in $V_{d1}$ with the variable density $d$ to generate a set of variables $S=\{\boldsymbol{x_{d1}^i} \in V_{d1}| i=1,2,\ldots,d^{N_{d1}}\}$;
  \item Map each $\boldsymbol{x_{d1}^i}$ in $S$ through $\boldsymbol{\mathcal{F}_T}$ to obtain the according $\boldsymbol{x_{d2}^i}$ and check out if it is in the workspace;
  \item Delete all the $\boldsymbol{x_{d1}^i}$ in $S$, to which the according $\boldsymbol{x_{d2}^i}$ is not in the workspace, to obtain the new set of sample variables $S'$.
\end{enumerate}

Comparing to $S$, $S'$ has less variables and then less samples to be calculated, which reduces the calculating time. In addition, feasible success rate of non-linear optimization for sample calculation is improved by deleting the invalid variables. Also, unrepresentative samples are removed. Consequently, the efficiency and quality of database generation can be significantly improved.

Besides, if the conditions below are satisfied,
\begin{enumerate}
  \item $V_{d2}$ is known;
  \item The inverse mapping of $\boldsymbol{\mathcal{F}_T}$ exists and is injective mapping;
  \item $N_{d1}>N_{d2}$
\end{enumerate}

Then we can select sample variables as follow:
\begin{enumerate}
  \item Randomly or evenly select multiple vectors in $V_{d2}$ with the variable density $d$ to generate a set of variables $S=\{\boldsymbol{x_{d2}^i} \in V_{d2}| i=1,2,\ldots,d^{N_{d2}}\}$;
  \item Map each $\boldsymbol{x_{d2}^i}$ through the inverse mapping of $\boldsymbol{\mathcal{F}_T}$ to obtain the corresponding $\boldsymbol{x_{d1}^i}$ and then the sample variable set $S=\{\boldsymbol{x_{d1}^i} \in V_{d1}| i=1,2,\ldots,d^{N_{d2}}\}$.
\end{enumerate}

As $N_{d1}>N_{d2}$, the number of samples to be calculated is reduced with the same variable density by this method. In addition, similar to Section \ref{sec:database-mapm}, more probable sizes can be chosen and more reasonable distribution can be obtained.

In practice, we can select the suitable method according to the specific situation.

\subsection{Database Generation}    \label{sec:database-flow}
The procedure of database generation based on dimension-reducing mapping is as follow :

\begin{enumerate}
  \item Classify the components of the input variable $\boldsymbol{X}$. The high dimensional dependent components $\{\boldsymbol{x_{id}^i}| i=1,2,\ldots,n_M\}$ and the independent components need to be transformed $\{\boldsymbol{x_{d1}^j}| j=1,2,\ldots,n_T\}$ are extracted, where $n_M$ and $n_T$ represent the count of groups of $\boldsymbol{x_{id}}$ and $\boldsymbol{x_{d1}}$ in $\boldsymbol{X}$, that is, the count of merging mapping and transforming mapping need to be done, respectively. The remainder is denoted as $\boldsymbol{x_c}$. Then $\boldsymbol{X}$ can be denoted as
    \begin{equation}
    \boldsymbol{X}=\{\boldsymbol{{x_{id}}^1},\boldsymbol{{x_{id}}^2},\ldots,\boldsymbol{{x_{id}}^{n_M}},\boldsymbol{{x_{d1}}^1},\boldsymbol{{x_{d1}}^2},\ldots,\boldsymbol{{x_{d1}}^ {n_T}},\boldsymbol{x_c}\}.
    \end{equation}

\item Perform merging mapping for each $\boldsymbol{x_{id}^i}$ and transforming mapping for each $\boldsymbol{x_{d1}^j}$ in $\boldsymbol{X}$
\begin{equation}
\begin{split}
\boldsymbol{x_{d}^i}=\boldsymbol{\mathcal{F}_{Mi}} (\boldsymbol{x_{id}^i}), i=1,2,\ldots,n_M, \\
\boldsymbol{x_{d2}^j}=\boldsymbol{\mathcal{F}_{Tj}} (\boldsymbol{x_{d1}^j}), j=1,2,\ldots,n_T.
\end{split}
\end{equation}

    The new variable after dimension-reducing mapping is
    \begin{equation} \boldsymbol{X_L}=\{\boldsymbol{{x_d}^1},\boldsymbol{{x_d}^2},\ldots,\boldsymbol{{x_d}^{n_M}},\boldsymbol{{x_{d2}}^1},\boldsymbol{{x_{d2}}^2},\ldots,\boldsymbol{{x_{d2}}^ {n_T}},\boldsymbol{x_c}\}.
    \end{equation}

  \item If there is any new components in $\boldsymbol{X_L}$ can be further mapped, let $\boldsymbol{X}=\boldsymbol{X_L}$ and repeat (1) and (2). Otherwise, move to (4).

  \item According to Section \ref{sec:database-mapm} and \ref{sec:database-mapt}, $\boldsymbol{{x_d}^i}$ and $\boldsymbol{{x_{d1}}^j}$ are used to solve non-linear optimization. Construct sample variable
  \begin{equation} \boldsymbol{X_S}=\{\boldsymbol{{x_d}^1},\boldsymbol{{x_d}^2},\ldots,\boldsymbol{{x_d}^{n_M}},\boldsymbol{{x_{d1}}^1},\boldsymbol{{x_{d1}}^2},\ldots,\boldsymbol{{x_{d1}}^ {n_T}},\boldsymbol{x_c}\}.
  \end{equation}

    Select $N_D$ groups of sample variables by the method illustrated in Section \ref{sec:database-mapm} and \ref{sec:database-mapt} and generate a set of sample variables
    \begin{equation}
	S=\{\boldsymbol{{X_S}^i}| i=1,2,\ldots,N_D\}.
    \end{equation}

  \item Solve the non-linear optimization problem specified by each $\boldsymbol{{X_S}^i}$ in $S$ to obtain the optimal parameter
    \begin{equation}
	\boldsymbol{C^{\ast i}}=O(\boldsymbol{{X_S}^i}).
    \end{equation}

    The optimal database can be established
    \begin{equation}
	D=\{(\boldsymbol{{X_S}^i},\boldsymbol{C^{\ast i}})| i=1,2,\ldots,N_D \}.
    \end{equation}

\end{enumerate}


\section{On-line Learning Based on the Multivariable-Multiple-Regression}
\subsection{Regression and Prediction}
\subsubsection{Multivariable-Multiple-Regression}
Assume the regression model is
\begin{equation}
\boldsymbol{C}=R(D,\boldsymbol{X}),
\end{equation}
where $\boldsymbol{X} \in \mathbb{R}^{N_X}$ is the input variable and $\boldsymbol{C} \in \mathbb{R}^{N_C}$ is the output variable.
Traditional regression models generally can only build the relationship between the single-dimensional output and input variables. However, regressions models in LO are always required to output multiple-dimensional parameters. Thus, we construct the multivariable-multiple-regression model, where multivariable means multiple-dimensional inputs and multiple means multiple-dimensional outputs.

First, we build the regression model $R^{(i)}$ for each element of output variable $C^{(i)}$
\begin{equation}
C^{(i)}=R^{(i)}(D,\boldsymbol{X}).
\end{equation}

Then multivariable-multiple-regression model can be regarded as the set of all regression models
\begin{equation}
\boldsymbol{R}=\{R^{(i)} |i=1,2,\ldots,N_C\}.
\end{equation}

As each $R^{(i)}$ is built independently, the input variable for each $R^{(i)}$ can be different. In other words, the related components in $\boldsymbol{X}$ which have influence to $C^{(i)}$ may be different. Therefore, for each $R^{(i)}$, we can select specific components in $\boldsymbol{X}$ to serve as input variables, which are called feature vector, denoted as $\boldsymbol{X_F}$.

Then $R^{(i)}$ can be built independently regarding to $\boldsymbol{{X_F}^{(i)}}$
\begin{equation}
C^{(i)}=R^{(i)} (D,\boldsymbol{{X_F}^{(i)}}).
\end{equation}

\subsubsection{Regression Methods}
Different regression methods can be adopted to build maps between $\boldsymbol{X}$ and $\boldsymbol{C}$ by employing the database. Three methods, $k$-Nearest Neighbours Regression ($k$-NN) \cite{ref:cover67}, Support Vector Machine Regression (SVR) \cite{ref:wu04} and Gaussian Process Regression (GPR) \cite{ref:seeger04}, are mainly discussed in this paper, which is detailed in Section \ref{sec:result-regression}.

\subsection{Parameter Refinement}
The parameter output by the regression and prediction $\boldsymbol{C'}$ may not satisfy the constraints, which requires a further refinement. In general, $\boldsymbol{C'}$ is taken as the initial parameter to solve the original optimization problem
\begin{equation}    \label{equ:refineo}
\boldsymbol{C^\ast}=O_{\boldsymbol{C'}}(\boldsymbol{X}).
\end{equation}

As $\boldsymbol{C'}$ is close to the global minimum, the local solution yield with the initial parameter $\boldsymbol{C'}$ is likely to be the global minimum, and the calculation time is largely reduced comparing to that with a randomly selected initial point.

However, when the objective function and constraints are very complex, Equation \ref{equ:refineo} may hardly be solved in real-time. To further reduce the calculation time, another way is to simplify the objective function, such as remove the objective function to perform a non-objective optimization
\begin{equation}    \label{equ:refinen}
\boldsymbol{C^\ast}=\bar{O}_{\boldsymbol{C'}}(\boldsymbol{X}).
\end{equation}

This method is effective for improving calculation speed, but only can output near-optimal solutions, rather than the accurate optimum. Unfortunately, there is no guarantee for the quality of the solutions by this method. When $F(\boldsymbol{C})$ changes sharply around $\boldsymbol{C^\ast}$, very small offset of $\boldsymbol{C}$ may lead to a large variation of $F$, which results in the very high cost increase rate.

Assume $F(\boldsymbol{C})$ is differentiable around the global minimum $\boldsymbol{C^\ast}$, the total differential of $F(\boldsymbol{C})$ at $\boldsymbol{C^\ast}$ is
\begin{equation}
dF(\boldsymbol{C^\ast})=\sum_{i=1}^{N_C} \frac{\partial{F}}{{\partial{C}}_i}(\boldsymbol{C^\ast}){dC}_i,
\end{equation}
which indicates the changing rate of $F(\boldsymbol{C})$ around $\boldsymbol{C^\ast}$.

The relative changing rate of the objective function around $\boldsymbol{C^\ast}$ is defined as
\begin{equation}
{dF}_r (\boldsymbol{C^\ast})=\frac {dF(\boldsymbol{C^\ast})}{F(\boldsymbol{C^\ast})}.
\end{equation}

When ${dF}_r(\boldsymbol{C^\ast})$ is small, the cost increase rate is low with the little parameter deviation $d\boldsymbol{C}$. On the contrary, when ${dF}_r(\boldsymbol{C^\ast})$ is large, the cost increase rate may be very high with a very little $d\boldsymbol{C}$.

Therefore, we take the refinement method as an unfixed sub-method of the LO model $\mathcal{L}$. In practice, refinement method is chosen according to the requirement and situation in the specific application. For example, we can choose Equation \ref{equ:refinen} under the situations which are sensitive to computation speed but require little of the optimality, and choose Equation \ref{equ:refineo} when it can run in real-time.

\section{Case: Kinematical Based Trajectory Planning for Haptic Feedback Manipulators}
\subsection{Continuous Trajectory Planning of Haptic Feedback Systems}  \label{sec:case-HFS}
In virtual reality systems, the real world is simulated by constructing a virtual environment, which provides visual sense to users. However, the corresponding haptic feedback is hardly achieved when users tend to touch the objects in the virtual environment. The missing of haptic feedback significantly limits the performance of interaction and immersion, especially in virtual manipulation systems.

\begin{figure}[h]
 \centering
 \includegraphics[width=0.8\columnwidth]{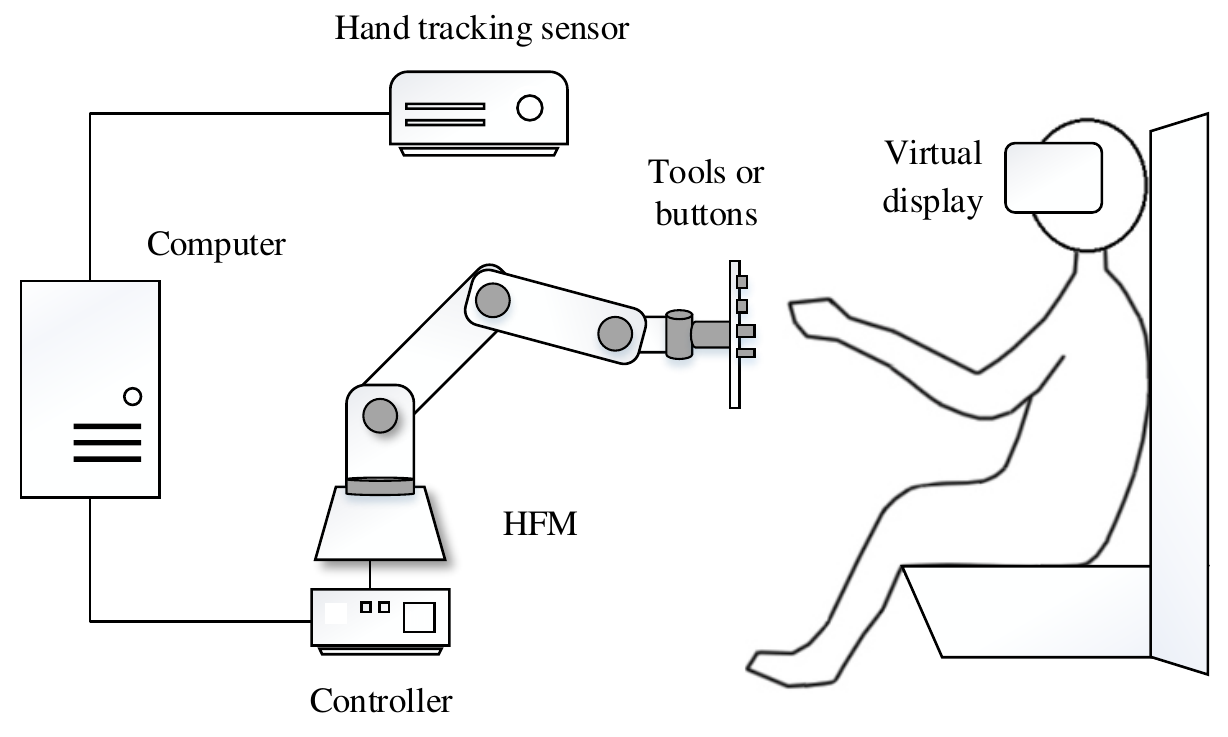}
 \caption{Structure of the haptic feedback system}
 \label{fig:HFSstru}
\end{figure}

\begin{figure}[h]
 \centering
 \includegraphics[width=0.8\columnwidth]{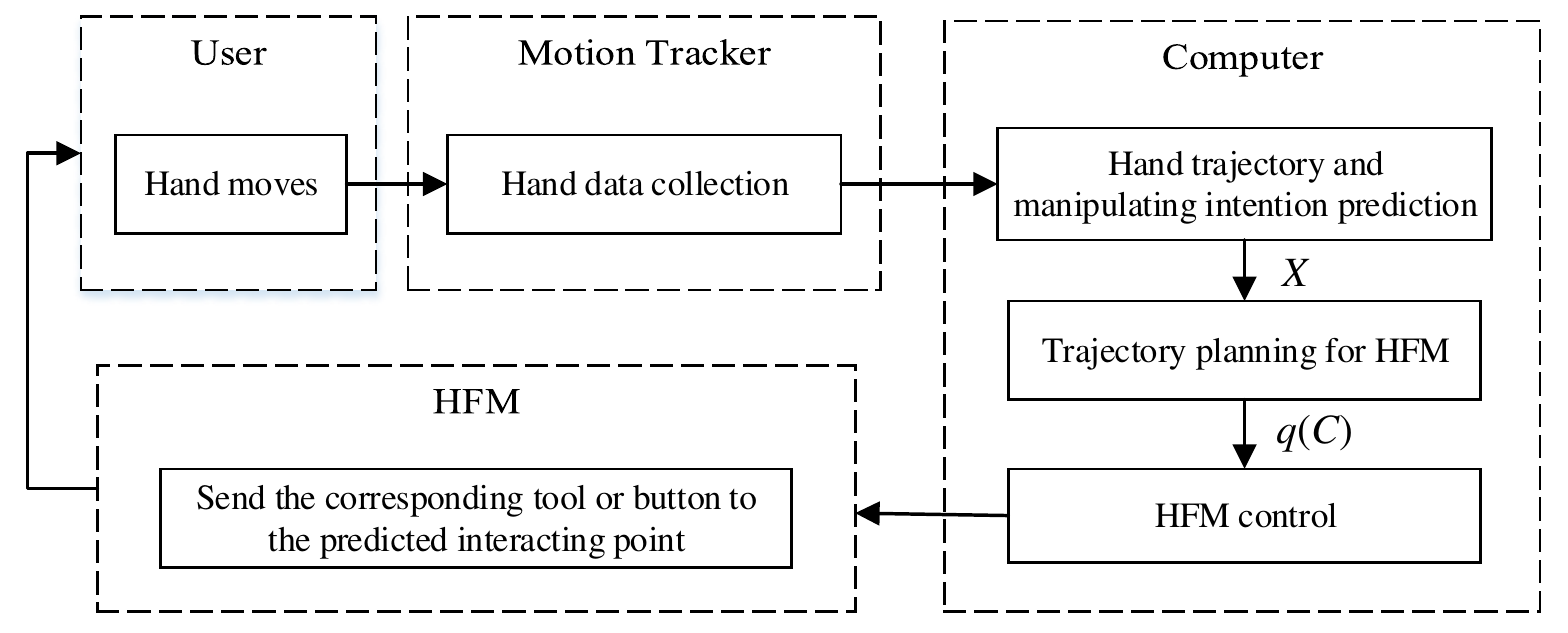}
 \caption{Workflow of the haptic feedback system}
 \label{fig:HFSflow}
\end{figure}

To achieve haptic feedback in virtual manipulation systems, a haptic feedback system based on the servo serial manipulator is constructed, shown as Figure \ref{fig:HFSstru}. A serial manipulator is served as the haptic feedback manipulator (HFM), with diverse kinds of tools or buttons linked with the end-effector, which is placed before the user to provide the tactile sensations. As user's hand moves, the motion data of the hand are collected by the hand motion trackers and then processed by the computers to predict the future trajectory of the hand and the manipulating intention, and then determine the interacting time $t_c$ and the interacting position $\boldsymbol{p_c}$, in which human hand and the specific tool or button contact. Then trajectory planning and control for HFM are performed to send the tool or button to $\boldsymbol{p_c}$ at $t_c$. As the hand moves, the predicted hand trajectory and interacting point are updated continuously, which requires the continuous re-planning for the HFM. When user's hand finally reach the actual objective position and conduct manipulation, the user is able to see the virtual hand operating the tool or button in virtual environment through the virtual display, while the corresponding tool or button reaches the objective position to provide haptic feedback at the same time. The workflow is shown in Figure \ref{fig:HFSflow}.

In haptic feedback applications, the HFM and the user work in a shared workspace, and direct contacts between the end-effector of HFM and user's hand are performed, which requires safety measures for human-robot interaction. To avoid injury from robots to humans, the motion ranges of the HFM and user's hand are restricted. First, the reachable workspace of the HFM is divided into the interacting area $W_C$, the effective workspace $W_E$ and the prohibiting area $W_P$, which satisfy
\begin{equation}
W_C \subset W_E
\end{equation}
\begin{equation}
W_E+W_P=W
\end{equation}

Then the motion range of the HFM and the hand are restricted to the side of $W_E$ and $W_P$, respectively. And all the contacts are performed in $W_C$.

The trajectory of the HFM is continuous re-planed during the whole haptic feedback process. Taking $T_p$ as the period to update the interacting point, a sequence of interacting points $(\boldsymbol{{p_c}^0},\boldsymbol{{p_c}^1},\ldots,\boldsymbol{{p_c}^n)}\subset S_E$ is obtained. As the hand moves, the predicted interacting point slightly changes and gradually converges to the real interacting point. In the $i$th period, the interacting configuration according to $\boldsymbol{{p_c}^i}$ is taken as the objective configuration of trajectory planning, denoted as $\boldsymbol{q_c} \in \mathbb{R}^{N_J}$. For sake of motion continuity, the current configuration and velocity are taken as the initial configuration and velocity of the trajectory planning, denoted as $\boldsymbol{q_0} \in \mathbb{R}^{N_J}$ and $\boldsymbol{\omega_0} \in \mathbb{R}^{N_J}$, respectively. Then the input of the trajectory planning problem can be specified as $\boldsymbol{X}=(\boldsymbol{q_0},\boldsymbol{q_c},\boldsymbol{\omega_0}) \in \mathbb{R}^{3N_J}$.

\subsection{Trajectory Planning Based on Non-linear Optimization}
\subsubsection{Trajectory Parameterization}
Considering the kinematical trajectory planning for the HFM, the trajectory is encoded by the trapezoidal velocity profile, which consists of three phases: the uniform acceleration phases $(0-t_1)$, the maximum velocity phase $(t_1-t_2)$ and the uniform deceleration phase $(t_2-t_f)$. The acceleration and deceleration phase have the same acceleration $a$. And the velocity for maximum velocity phase is $\omega_m$. For each joint, the initial joint position $q_0$, the objective joint position $q_c$ and the initial velocity $\omega_0$ are known. There are five variables $a$, $\omega_m$, $t_1$, $t_2$ and $t_f$, which are linked by three equations:
\begin{equation}\label{equ:trap1}
\frac{1}{2} \omega_m (t_f+t_2-t_1) + \frac{1}{2} \omega_0 t_1 = q_c-q_0,
\end{equation}
\begin{equation}\label{equ:trap2}
\omega_m=\omega_0+at_1,
\end{equation}
\begin{equation}\label{equ:trap3}
\frac{\omega_m-\omega_0}{t_1} =\frac{\omega_m}{t_f-t_2}.
\end{equation}

Thus, there are only two independent variables while the others can be derived by Equation \ref{equ:trap1}-\ref{equ:trap3}.

Define
\begin{equation}
\hat{r}(t,t_1,t_2)=r(t_1-t)-r(t-t_2),
\end{equation}
where $r(t)$ is the step function
\begin{equation}
r(t)=\left\{ \begin{array}{ll}
1 & t \ge 0 \\
0 & t<0.
\end{array} \right .
\end{equation}

Then the trajectory of each joint is represented as
\begin{equation}
\ddot{q}(t)=a \hat{r}(t,t_1,t_2 ),
\end{equation}
\begin{equation}
\dot{q}(t)=\omega_0+a \hat{r}(t,t_1,t_2)t,
\end{equation}
\begin{equation}
q(t)=q_0+\omega_0 t + \frac{1}{2} a \hat{r}(t,t_1,t_2) t^2.
\end{equation}

\subsubsection{Non-Linear optimization} \label{sec:case-opt}
\quad

\textbf{A. Optimization}

A $N_J$-DOF serial robot is adopted as the HFM, and the initial configuration, the objective configuration and the initial velocity are known as $\boldsymbol{q_0} \in \mathbb{R}^{N_J}$, $\boldsymbol{q_c} \in \mathbb{R}^{N_J}$, $\boldsymbol{\omega_0} \in \mathbb{R}^{N_J}$, respectively. Then the input of the trajectory planning based on non-linear optimization is $\boldsymbol{X}=(\boldsymbol{q_0},\boldsymbol{q_c},\boldsymbol{\omega_0}) \in \mathbb{R}^{3N_J}$.

For each joint, $\omega_m$ and $t_f$ are selected as the independent variables, and other variables can be derived by Equation \ref{equ:trap1}-\ref{equ:trap3}.

All the joints are required to move simultaneously, that is, have the same motion time:
\begin{equation}
{t_f}^1={t_f}^2=\ldots={t_f}^{N_J}=t_f.
\end{equation}

The velocity for maximum velocity phase of the $i$th joint is $\omega_{m,i}$. Thus, trajectory planning for the $N_J$-DOFs HFM have $N_C=N_J+1$ parameters for optimization
\begin{equation}
C=(\omega_{m,1},\omega_{m,2},\ldots,\omega_{m,N_J},t_f) \in \mathbb{R}^{N_J+1}.
\end{equation}

Then the optimization problem describes in Equation \ref{equ:optim} can be specified as
\begin{equation}
\boldsymbol{C^\ast}=O(\boldsymbol{X}).
\end{equation}

\textbf{B. Objective function}

One the one hand, to ensure the safety, the HFM is required to move softly with the accelerations as small as possible. On the other hand, to achieve the rapidity of motion, the HFM is required to reach the objective configuration quickly with the motion time as small as possible. There is a conflict between safety and rapidity. Thus, to compromise the two aspects, the objective function is designed as a balance between acceleration and motion time:
\begin{equation}
F(\boldsymbol{C})=\mu(\frac{t_f}{t_{max}})^2+\frac{1-\mu}{N_J} \sum_{i=1}^{N_J} {(\frac{a_i}{a_{i,max}})}^2
\end{equation}
where $t_{max}$ is the maximum allowable motion time, $a_{i,max}$ is the maximum allowable acceleration of the $i$th joint, and $\mu$ is the weighting coefficient representing the weighting of acceleration and motion time, which satisfies $0 \le \mu \le 1$.

\textbf{C. Constraints}

When the HFM works, the velocity and the acceleration should not exceed the mechanism limitation:
\begin{equation}
0<\omega_{m,i}<\dot{q}_{i,max}£¬i=1,\ldots,N_J,
\end{equation}
\begin{equation}
0<a_i<\ddot{q}_{i,max}£¬i=1,\ldots,N_J.
\end{equation}

In addition, to ensure the safety of human-robot interaction, the HFM is required to reach the interaction configuration before the hand reach the interacting point so that the HFM is static when contacts with user's hand. Thus, the motion time of the HFM should not exceed the interacting time $t_c$:
\begin{equation}
t_f<t_c.
\end{equation}

Besides, the parameters should satisfy the principle of the trapezoidal velocity profile:
\begin{equation}
0 \le t_1 \le t_f/2,
\end{equation}
\begin{equation}
t_1 \le t_2 \le t_f.
\end{equation}

\subsection{Database Generation Based on Dimension-Reducing Mapping}    \label{sec:case-database}
According to Section \ref{sec:database-flow}, the elements of the input variable $\boldsymbol{X}=(\boldsymbol{q_0},\boldsymbol{q_c},\boldsymbol{\omega_0})$ are classified into different types and handled respectively. First, the optimization problem specified in Section \ref{sec:case-opt} is related to the offset $\boldsymbol{q_c}-\boldsymbol{q_0}$ rather than the individual $\boldsymbol{q_c}$ and $\boldsymbol{q_0}$. Therefore, $(\boldsymbol{q_0},\boldsymbol{q_c})$ is a dependent component and can be merged
\begin{equation}    \label{equ:casemap1}
\begin{split}
\boldsymbol{q_f}=\boldsymbol{\mathcal{F}_M}(\boldsymbol{q_0},\boldsymbol{q_c})=\boldsymbol{q_c}-\boldsymbol{q_0}    \\
\boldsymbol{\mathcal{F}_M}:(V_q,V_q) \subset \mathbb{R}^{2N_J} \to V_{q_f} \subset \mathbb{R}^{N_J}
\end{split}
\end{equation}
where $V_q$ and $V_{q_f}$ are the allowable motion range and the allowable offset range of the HFM, respectively.

Then the sample variable can be selected as $\boldsymbol{X_S}=(\boldsymbol{q_f},\boldsymbol{\omega_0})$, as $\boldsymbol{q_f}$ and $\boldsymbol{\omega_0}$ are independent variables.
In addition, according to Section \ref{sec:case-HFS}, the sample variables should be in the workspace of the HFM, that is, the initial and objective position of the end-effector should be located in the effective workspace and the interacting area, respectively:
\begin{equation}
\begin{split}
\boldsymbol{p_0} \in W_E,    \\
\boldsymbol{p_c} \in W_C.
\end{split}
\end{equation}

Thus, to determine whether an input variable is in the work range of the HFM, $\boldsymbol{q_0}$ and $\boldsymbol{q_c}$ are transformed into the task space by the forward kinematics:
\begin{equation}    \label{equ:casemap2}
\begin{split}
\boldsymbol{p_0}=\boldsymbol{\mathcal{F}_{TFK}}(\boldsymbol{q_0}),     \\
\boldsymbol{\mathcal{F}_{TFK}}:V_q \subset \mathbb{R}^{N_J} \to W \subset \mathbb{R}^{N_J},	\\
\boldsymbol{p_c}=\boldsymbol{\mathcal{F}_{TFK}}(\boldsymbol{q_c}),     \\
\boldsymbol{\mathcal{F}_{TFK}}:V_q \subset \mathbb{R}^{N_J} \to W \subset \mathbb{R}^{N_J}.
\end{split}
\end{equation}

A set of sample variables $\boldsymbol{X_S}=(\boldsymbol{q_f},\boldsymbol{\omega_0})$ need to be selected to establish the database. The combinations of $\boldsymbol{q_f}$ and $\boldsymbol{\omega_0}$ are selected according to the practical motion curve to guarantee that the selected sample variables are more reasonable. In the haptic feedback process, the HFM starts from the static state and continuous be re-planned according to some intermediate points. Thus, we first consider $\boldsymbol{\omega_0}=\boldsymbol{0}$, and select multiple $\boldsymbol{q_f}$, to solve $\boldsymbol{C}=O(\boldsymbol{q_f},\boldsymbol{0})$ and obtain the parameter $\boldsymbol{C}=(\boldsymbol{\omega_m},t_f)$ and the corresponding trapezoidal velocity curve $\boldsymbol{\dot{q}}(\boldsymbol{C},t)$. Then some intermediate points of each curve are selected
\begin{equation}
t'=\alpha t_f,
\end{equation}
where $\alpha \in V_\alpha=[0,1)$. Then the new initial velocity $\boldsymbol{\omega_0'}$ and offset $\boldsymbol{q_f'}$ can be generated to serve as the sample variables
\begin{equation}
\boldsymbol{\omega_0'}=\boldsymbol{\dot{q}}(\boldsymbol{C},t'),
\end{equation}
\begin{equation}
\boldsymbol{q_f'}=\boldsymbol{q}(t_f)-\boldsymbol{q}(t')-\boldsymbol{q_0}.
\end{equation}

The above process can be regarded as another transforming mapping
\begin{equation}    \label{equ:casemap3}
\begin{split}
(\boldsymbol{q_f},\alpha)=\boldsymbol{\mathcal{F}_T} (\boldsymbol{q_f'},\omega_0'),    \\
\boldsymbol{\mathcal{F}_T}:(V_{q_f},V_\omega) \subset \mathbb{R}^{2N_J} \to (V_{q_f},V_\alpha) \subset \mathbb{R}^{N_J+1}, \\
\end{split}
\end{equation}
where $V_\omega$ is the range of the joint velocity.

From the above, according to Equation \ref{equ:casemap1}, \ref{equ:casemap2} and \ref{equ:casemap3}, the final dimension-reducing mapping of the input variables is
\begin{equation}
\begin{split}
\boldsymbol{X_L}=(\boldsymbol{q_f},\alpha)=\boldsymbol{\mathcal{F}}(\boldsymbol{q_0},\boldsymbol{q_c},\boldsymbol{\omega_0}), \\
\boldsymbol{\mathcal{F}}: V_X \subset \mathbb{R}^{3N_J} \to V_{X_L} \subset \mathbb{R}^{N_J+1}.
\end{split}
\end{equation}

Randomly or evenly choosing $N_1$ $\boldsymbol{q_f}$ and $N_2$ $\alpha$, $N_1 \ast N_2$ sample variables $\boldsymbol{X_S}=(\boldsymbol{q_f},\boldsymbol{\omega_0})$ can be obtained. Finally, database including $N_1 \ast N_2$ samples can be built by solving global optimization problems according to all $\boldsymbol{X_S}$s.

\subsection{Feature Selection for Multivariable-Multiple-Regression}    \label{sec:case-feature}
In multivariate-multiple-regression, we need to determine the feature vector for each element of $\boldsymbol{C}$. Two kinds of feature is proposed. One is to take $\boldsymbol{X}$ as the feature for each $C^{(i)}$:
\begin{equation}
t_f=f_t(\boldsymbol{X}),
\end{equation}
\begin{equation}
{\omega_m}^{(i)}={f_\omega}^{(i)}(\boldsymbol{X})  (i=1,2,\ldots,N_J).
\end{equation}

This method is simple, but the efficiency for building regression model and the accuracy of regression may be lower due to the effect of the irrelevant elements in $\boldsymbol{X}$.

The second way is to construct the individual feature for each $C^{(i)}$:
\begin{equation}
t_f=f_t(\boldsymbol{X}),
\end{equation}
\begin{equation}
{\omega_m}^{(i)}={f_\omega}^{(i)}({q_f}^{(i)},{\omega_0}^{(i)},t_f)  (i=1,2,\ldots,N_J).
\end{equation}

As all the joints are required to reach the objective position at the same time, the motion time $t_f$ is affected by all the elements in $\boldsymbol{X}$. Thus, the regression model for $t_f$ takes $\boldsymbol{X}$ as the feature. On the other hand, the maximum velocity for $i$th joint ${\omega_m}^{(i)}$ is only related to the offset and initial velocity of the corresponding joint and the motion time, which are taken as the feature for the regression model of ${\omega_m}^{(i)}$.

\section{Results}
\subsection{Database Generation by Dimension-Reducing Mapping}
We choose a 3-DOF HFM with the link parameters and motion limits shown in Table \ref{table:HFMpara} and Table \ref{table:HFMlimit}, respectively.

\begin{table}[h]
\centering
\caption{Link Parameters of the HFM}
\begin{tabular}{lcccc}
\hline\noalign{\smallskip}
Link & $\alpha$ /rad & $a$ /m & $\theta$ /rad & $d$ /m \\
\noalign{\smallskip}\hline\noalign{\smallskip}
L1   &   $-\pi/2$   & 0.2      & $q_1$ & 0      \\
L2   &   0          & 0        & $q_2$ & 0.3    \\
L3   &   0          & 0        & $q_3$ & 0.3    \\
\noalign{\smallskip}\hline
\end{tabular}
\label{table:HFMpara}
\end{table}

\begin{table}[h]
\centering
\caption{Motion Limits of the HFM}
\begin{tabular}{lcccc}
\hline\noalign{\smallskip}
Joint   & Motion Range ($^\circ$)  &Maximum velocity ($rad/s$) & Maximum acceleraton ($rad/s^2$) \\
\noalign{\smallskip}\hline\noalign{\smallskip}
Joint 1 &   $-\pi/2 \sim \pi/2$   & 1.75   &  15  \\
Joint 2 &   $-\pi/4 \sim \pi/4$   & 1.75   &  15  \\
Joint 3 &   $-\pi/4 \sim \pi/4$   & 1.75   &  15  \\
\noalign{\smallskip}\hline
\end{tabular}
\label{table:HFMlimit}
\end{table}

\begin{figure}
 \centering
 \includegraphics[width=0.4\columnwidth]{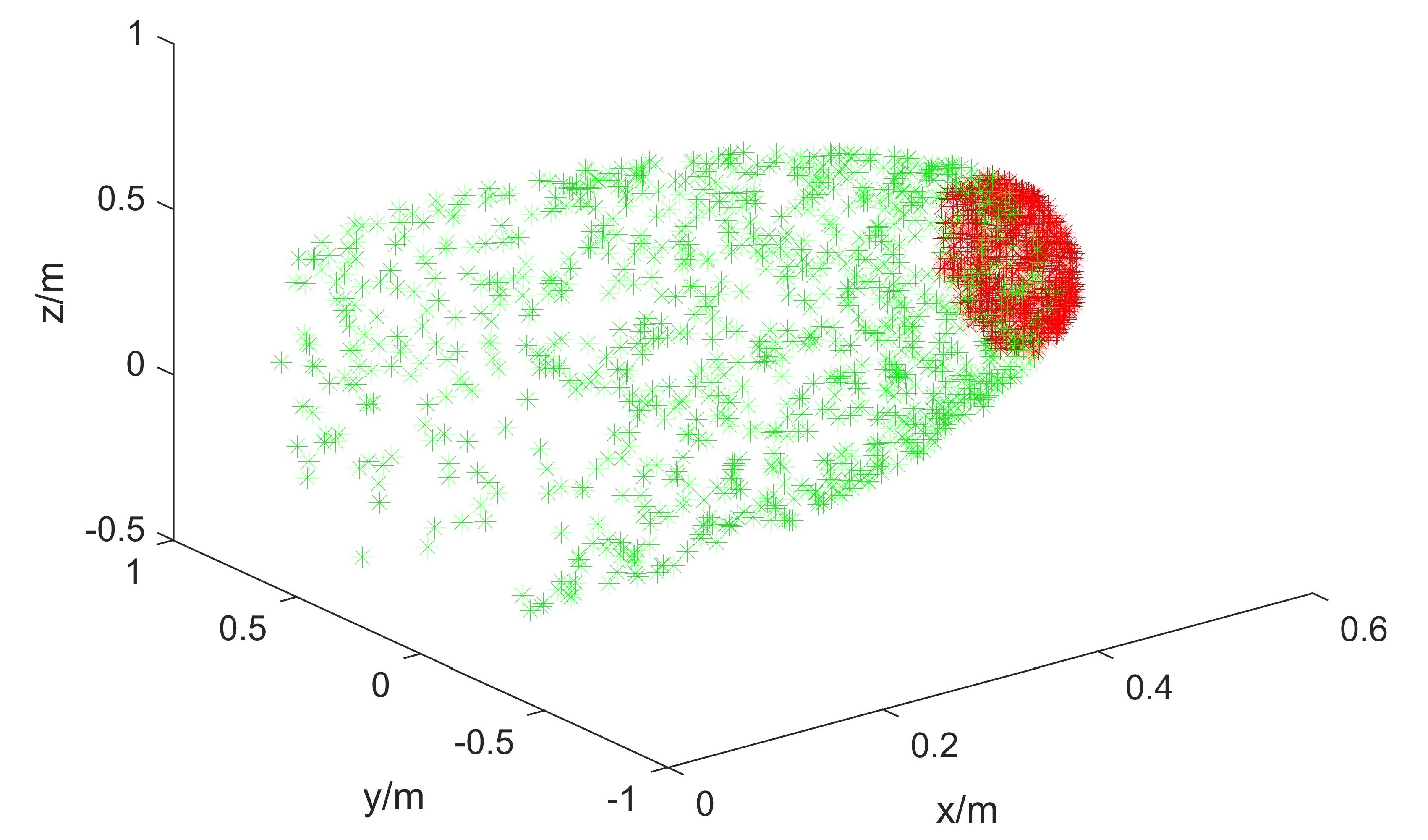}
 \caption{Workspace of the HFM}
 \label{fig:HFMworkspace}
\end{figure}

\begin{figure}
 \centering
 \includegraphics[width=0.4\columnwidth]{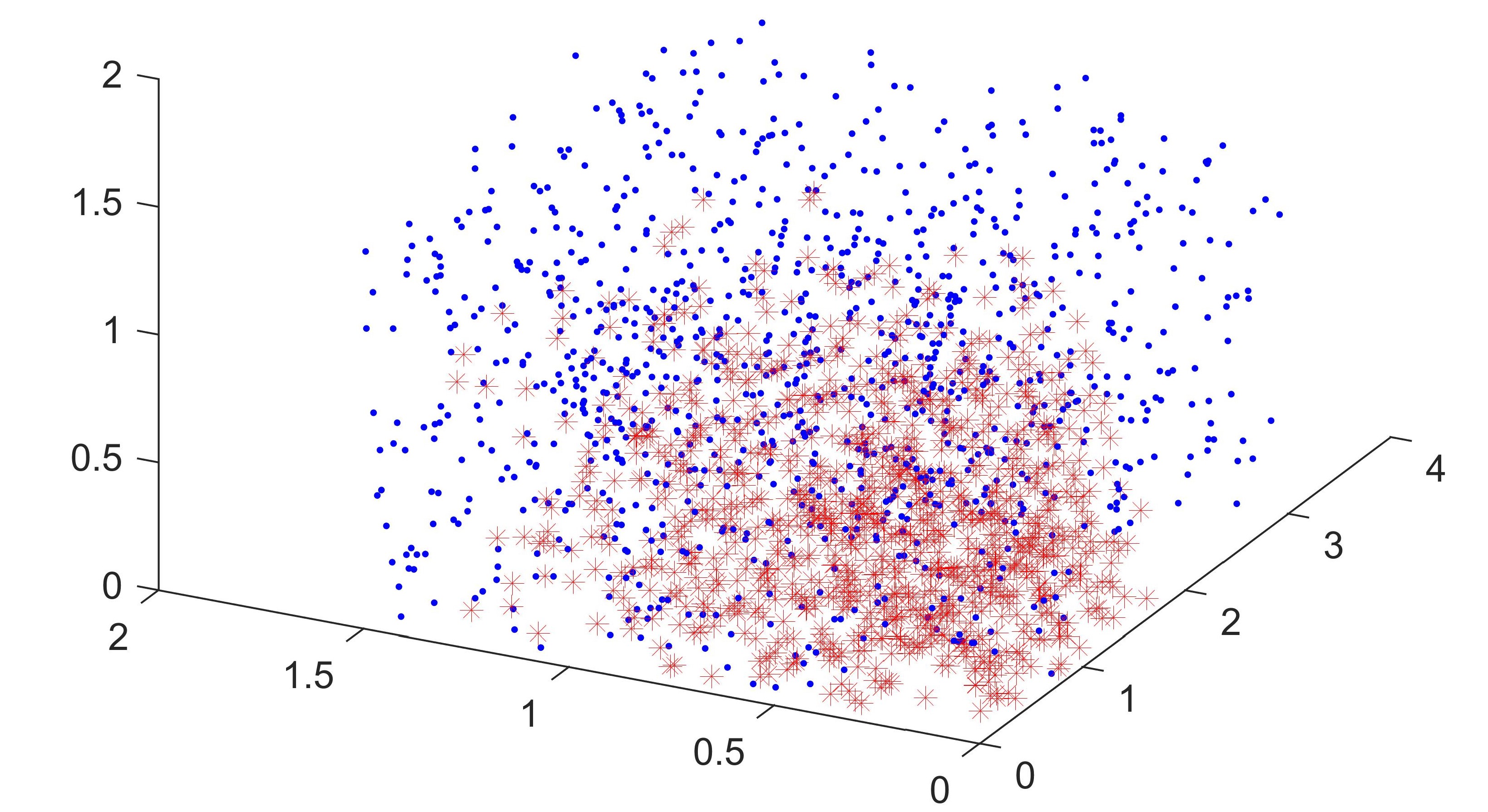}
 \caption{Comparison of of $\boldsymbol{q_f}$ selected by dimension-reducing and random method}
 \label{fig:HFMqf}
\end{figure}

The effective workspace $W_E$ and the interacting area $W_C$ are extracted, shown as the green and red points in Figure \ref{fig:HFMworkspace}, respectively. First, a set of $\boldsymbol{q_f}$ is generated by selecting $N_1=1000$ groups of $(\boldsymbol{p_0},\boldsymbol{p_c})$ which satisfied $\boldsymbol{p_0} \in W_E$ and $\boldsymbol{p_c} \in W_C$, and mapping them into the joint space, shown as the red points in Figure \ref{fig:HFMqf}. In contrast, another set of $\boldsymbol{q_f}$ are generated by randomly choosing 1000 groups of combination of joint variables in the motion range, shown as the blue points in Figure \ref{fig:HFMqf}. According to Figure \ref{fig:HFMqf}, the red points have a smaller coverage and a more concentrated distribution than the blue points, which indicates that the work range of the HFM does not cover the whole motion range of the robot. Therefore, transforming mapping is helpful to narrow the selection range of input variables, then generate more dense samples with the same amount and consequently improve the learning performance.

\begin{figure}
 \centering
 \includegraphics[width=0.4\columnwidth]{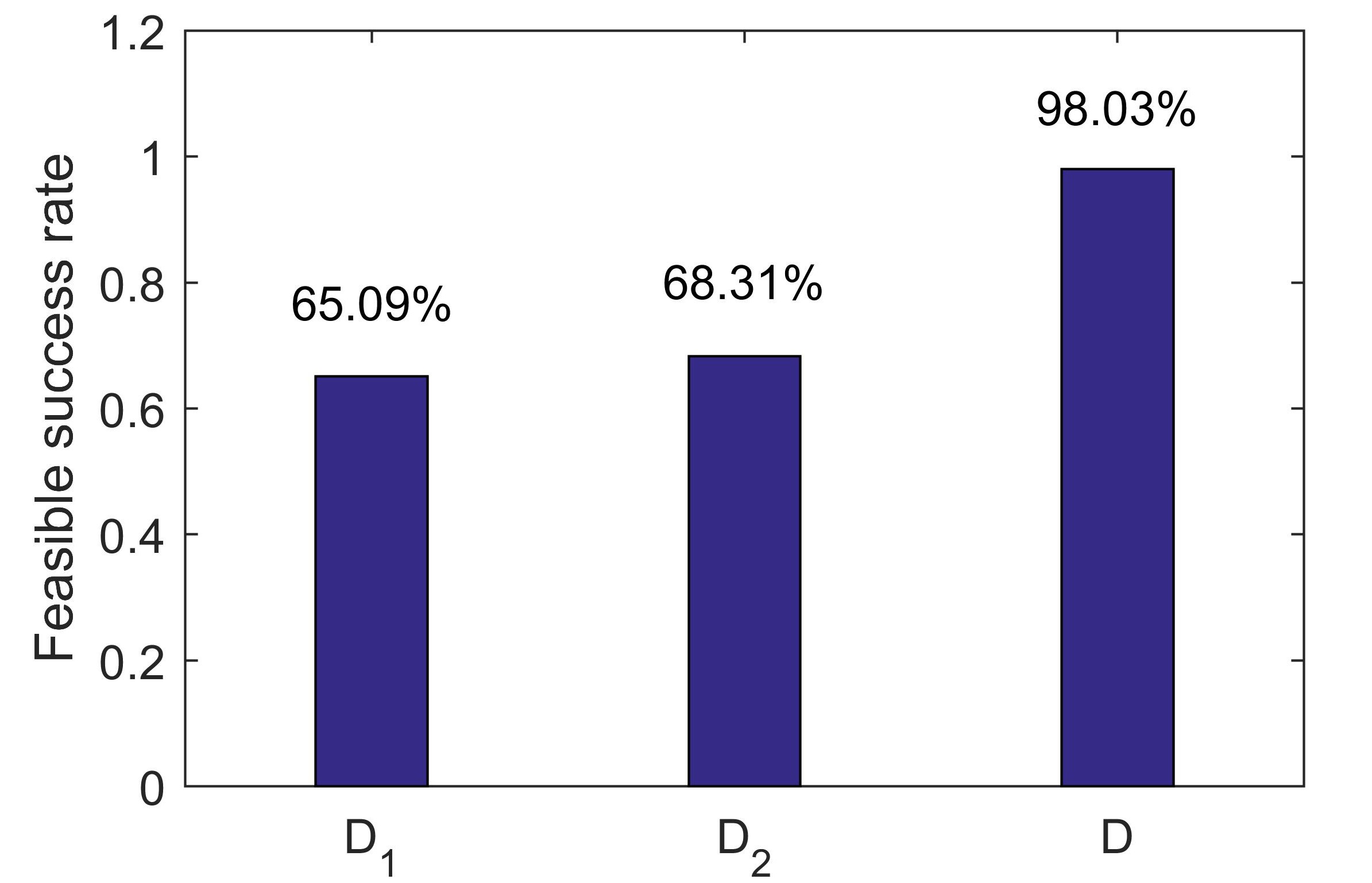}
 \caption{Feasible success rate for optimization in sample calculation with three methods}
 \label{fig:datafeasi}
\end{figure}

According to the sample selection method described in Section \ref{sec:case-database}, we choose $N_1=1000$ and $N_2=10$ and then obtain $N_1 \ast N_2=10000$ sample variables. Then database $D$ is generated by running global optimizations solved with multiple re-start SQP, in which the maximum re-start number ($N_{RO}$) is set to 10. On the other hand, two databases $D_1$ and $D_2$ are generated by randomly selecting sample variables with $d=5$ and solving optimizations by multiple re-start SQP with $N_{RO}=10$ and $N_{RO}=100$, respectively. The feasible success rate for optimization in sample calculation of the above three databases are shown in Figure \ref{fig:datafeasi}, which indicates that the success rate for sample calculation of the database generated by dimension-reducing mapping is much higher than traditional random method.

According to Figure \ref{fig:HFMqf} and \ref{fig:datafeasi}, dimension-reducing mapping of inputs in database generation can help to select more reasonable samples and avoid from invalid and unrepresentative samples, and increase the success rate for sample calculation, consequently improve the database quality.

\subsection{Feature Selection for Multivariable-Multiple-Regression}
To evaluate two features described in Section \ref{sec:case-feature}, we build the regression model of Gaussian-kernel-SVR and GPR, denoted as $R_{G-SVR}^1$, $R_{G-SVR}^2$, $R_{GPR}^1$ and $R_{GPR}^2$. We employ refinement with objective function and database consisting $N_D$ samples $D_{N_D}$ drawing from the original database $D$, and build LO models $\mathcal{L}(D_{N_D},R_{G-SVR}^1,O)$, $\mathcal{L}(D_{N_D},R_{G-SVR}^2,O)$, $\mathcal{L}(D_{N_D},R_{GPR}^1,O)$ and $\mathcal{L}(D_{N_D},R_{GPR}^2,O)$.

\begin{figure}
\centering
\subfloat{\includegraphics[width=0.25\textwidth]{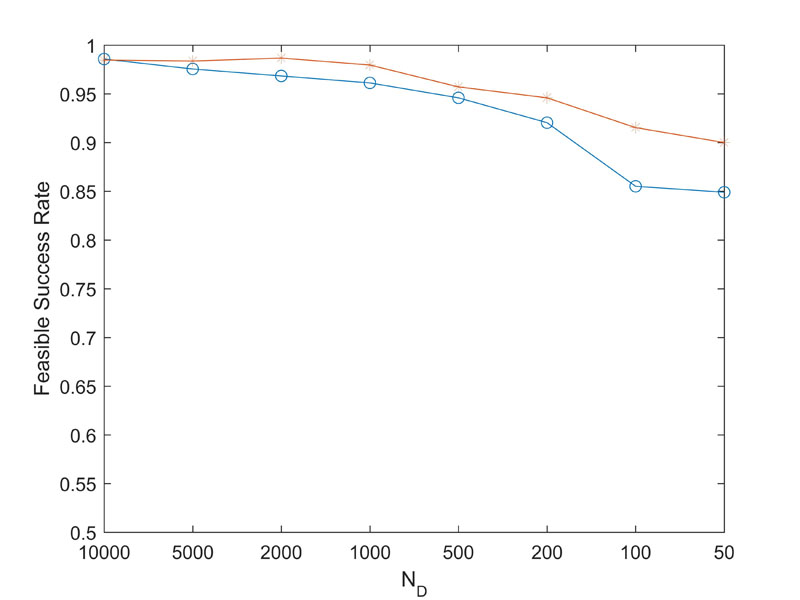}
\label{fig:rfeature-rs-svr}
}
\subfloat{\includegraphics[width=0.25\textwidth]{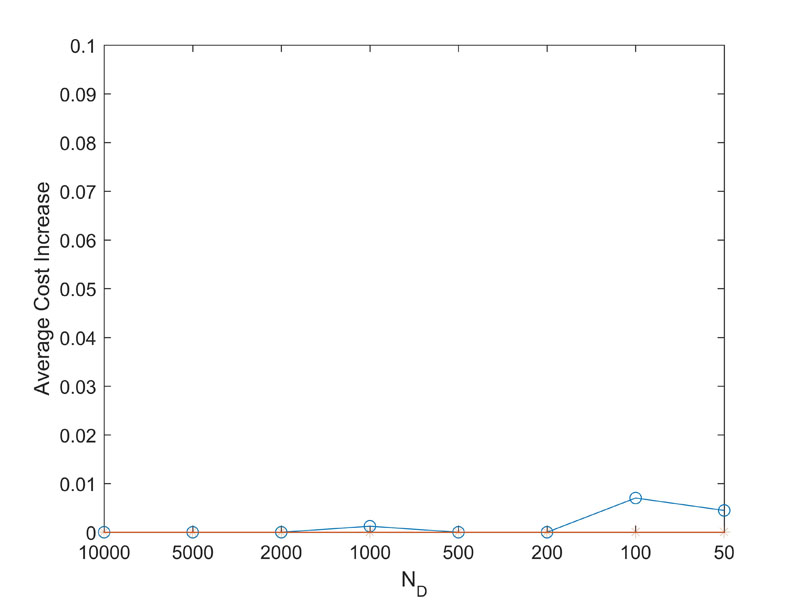}
\label{fig:rfeature-ef-svr}
}
\subfloat{\includegraphics[width=0.25\textwidth]{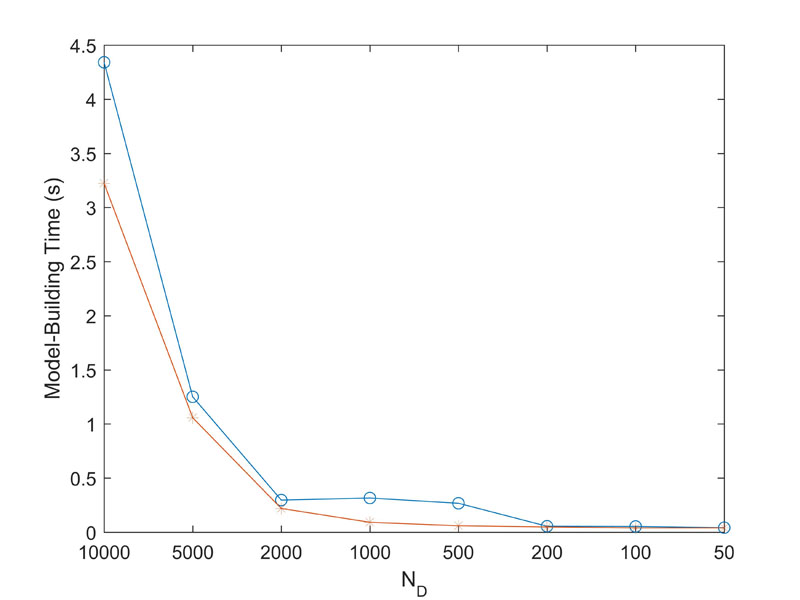}
\label{fig:rfeature-tb-svr}
}
\subfloat{\includegraphics[width=0.25\textwidth]{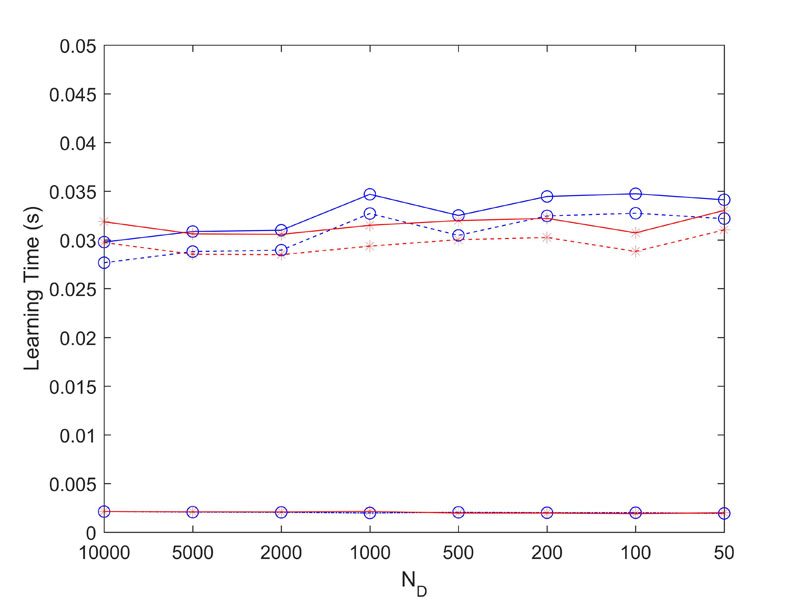}
\label{fig:rfeature-tl-svr}
}
\vspace{0.1pt}
\subfloat{\includegraphics[width=0.25\textwidth]{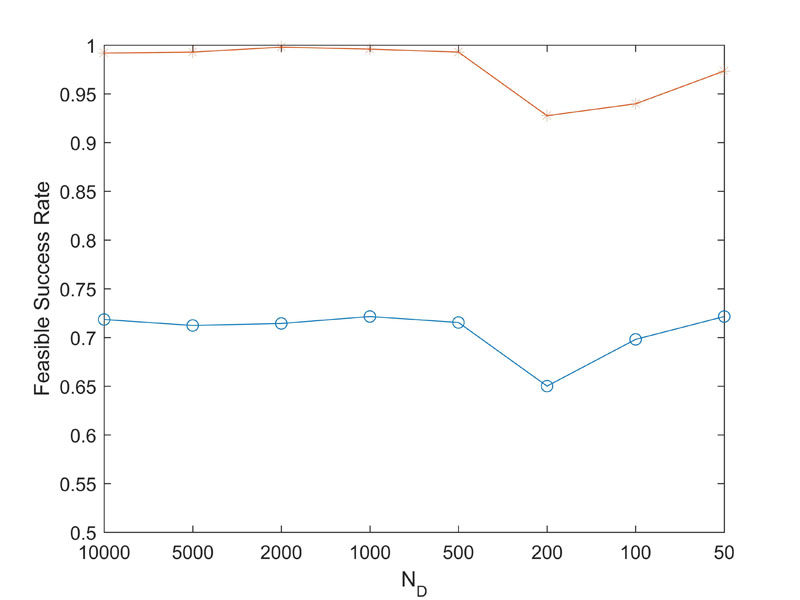}
\label{fig:rfeature-rs-gpr}
}
\subfloat{\includegraphics[width=0.25\textwidth]{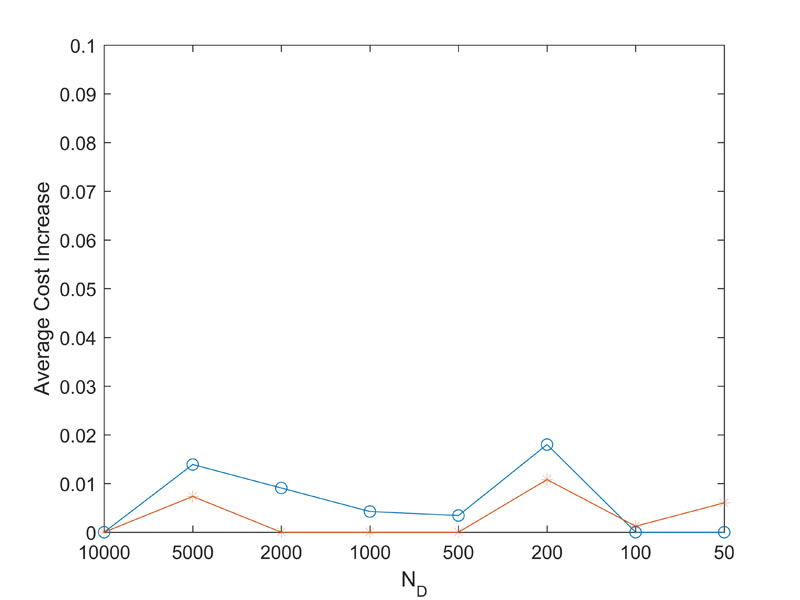}
\label{fig:rfeature-ef-gpr}
}
\subfloat{\includegraphics[width=0.25\textwidth]{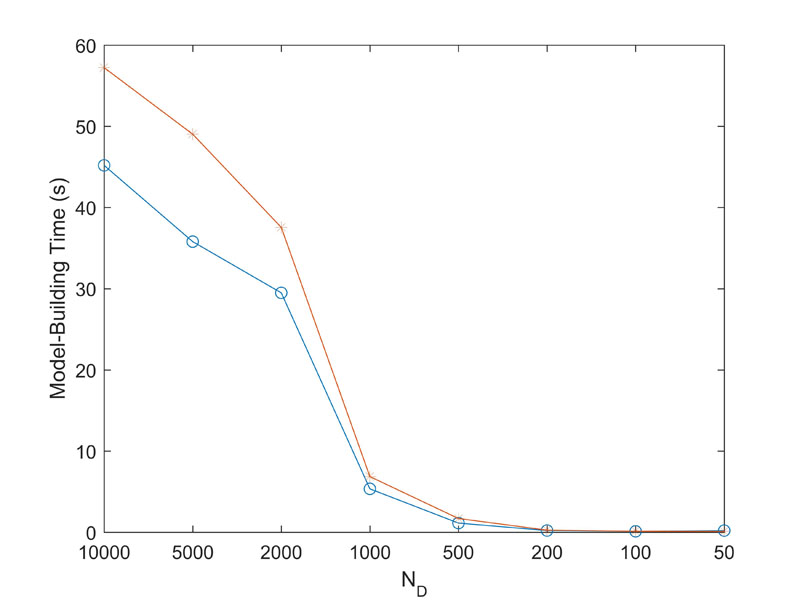}
\label{fig:rfeature-r-bsgpr}
}
\subfloat{\includegraphics[width=0.25\textwidth]{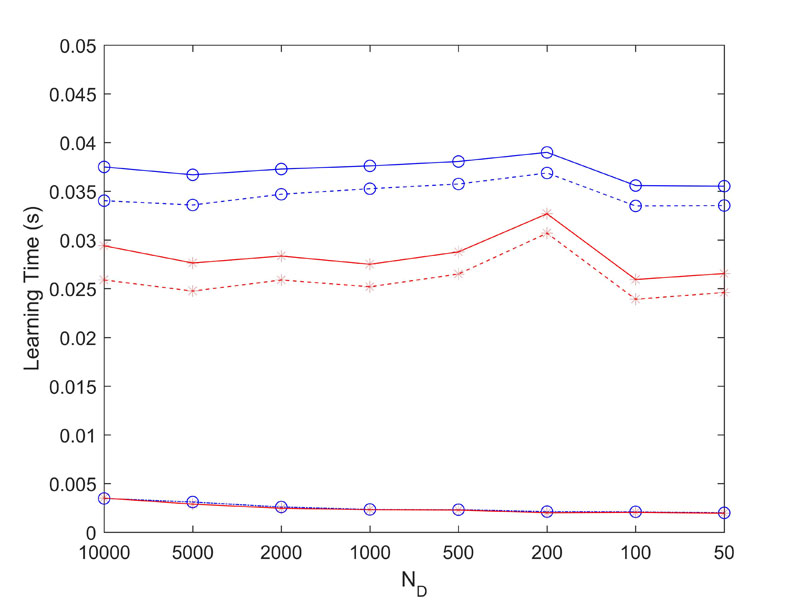}
\label{fig:rfeature-tl-gpr}
}
\caption{Learning results with different features (o: Feature 1, $\ast$: Feature 2; top: Gaussian-SVR, bottom: GPR; for the leaning time, dash-dotted: $T_{RP}$, dotted: $T_{RO}$; solid: $T_L$)}
\label{fig:rfeature}
\end{figure}

To evaluate the performance of the LO models, we generate a test set consisting of $N_r=1000$ samples by the dimension-reducing mapping method illustrated in Section \ref{sec:case-database}:
\begin{equation}
D_T=\{\boldsymbol{{X_S}^i}=(\boldsymbol{{q_f}^i},\boldsymbol{{\omega_0}^i}),\boldsymbol{{C^\ast}^i} |i=1,2,\ldots,N_r\}.
\end{equation}
Then we solve optimal parameters for each $\boldsymbol{X_S^i}$ in $D_T$ through LO models to test the performance, shown as Figure \ref{fig:rfeature}. With both Gaussian-SVR and GPR, the second feature derives higher feasible success rate, lower objective increase and smaller learning time than the first one, as the second one excludes effects of irrelevant elements of the input for each $R^{(i)}$.

\subsection{Regression Methods}     \label{sec:result-regression}
\begin{figure}
\centering
\subfloat[Feasible success rate]{\includegraphics[width=0.4\textwidth]{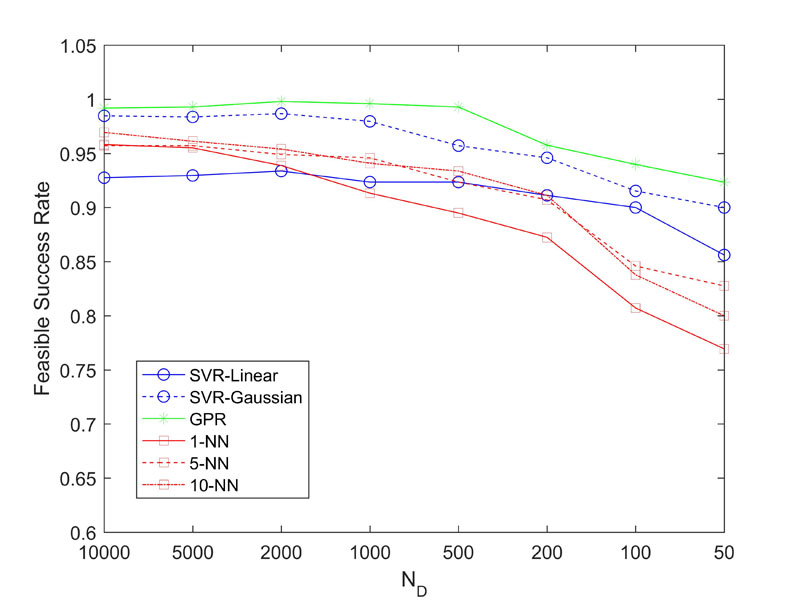}
\label{fig:rregression-rs}
}
\hspace{1pt}
\subfloat[Average cost increase]{\includegraphics[width=0.4\textwidth]{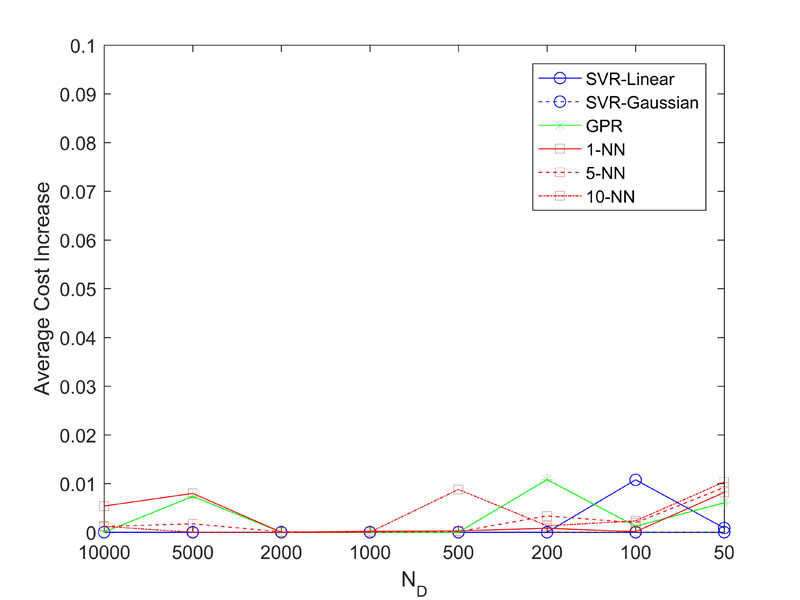}
\label{fig:rregression-ef}
}
\vspace{1pt}
\subfloat[Model-building time]{\includegraphics[width=0.4\textwidth]{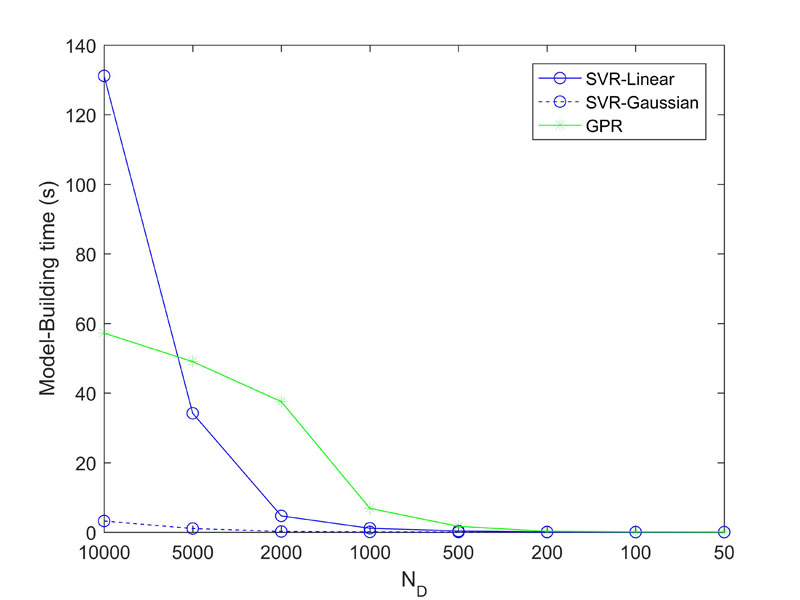}
\label{fig:rregression-tb}
}
\hspace{1pt}
\subfloat[On-line learning time]{\includegraphics[width=0.4\textwidth]{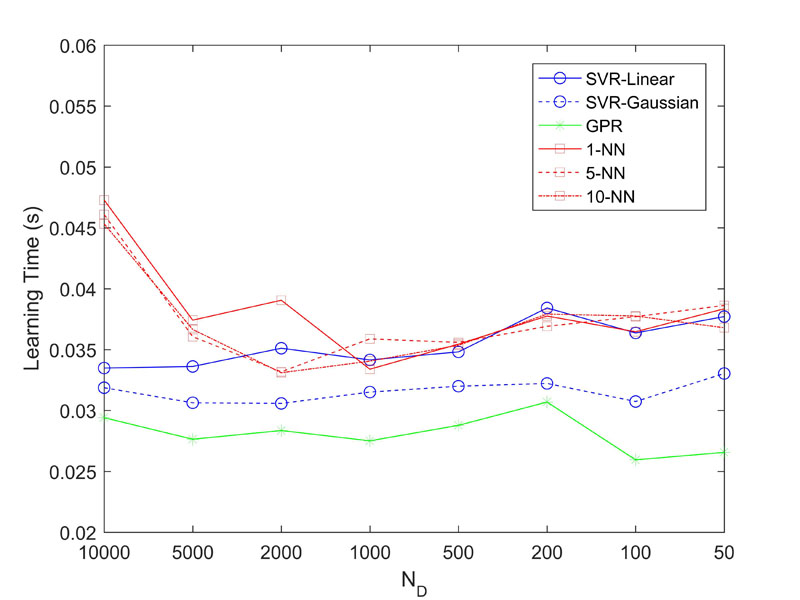}
\label{fig:rregression-tl}
}
\caption{Learning results with different regression methods}
\label{fig:rregression}
\end{figure}

\begin{figure}
\centering
\subfloat[Regression prediction time]{\includegraphics[width=0.4\textwidth]{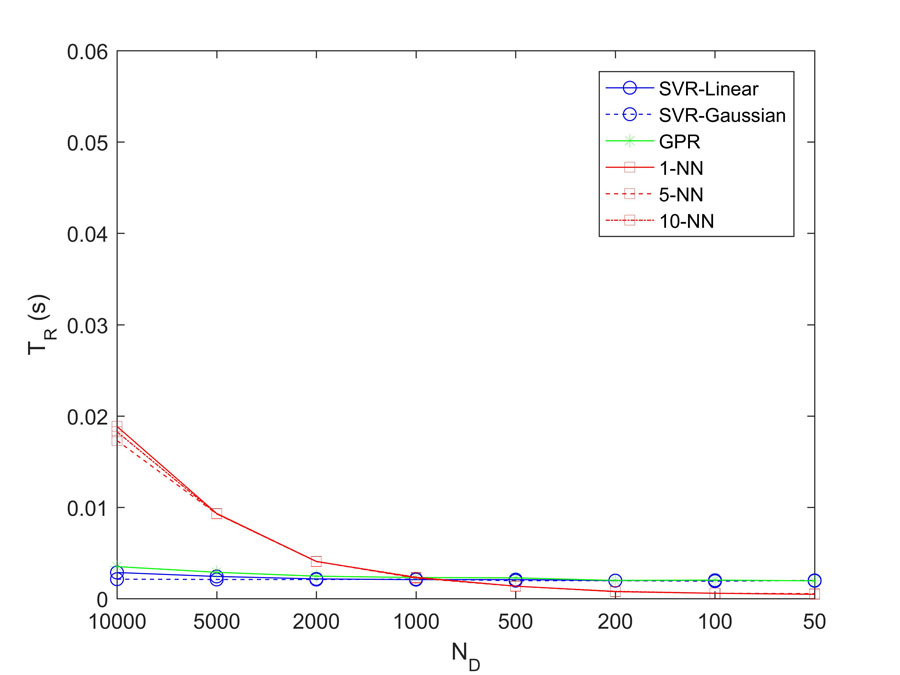}
\label{fig:rregression-tlr}
}
\hspace{1pt}
\subfloat[Refinement time]{\includegraphics[width=0.4\textwidth]{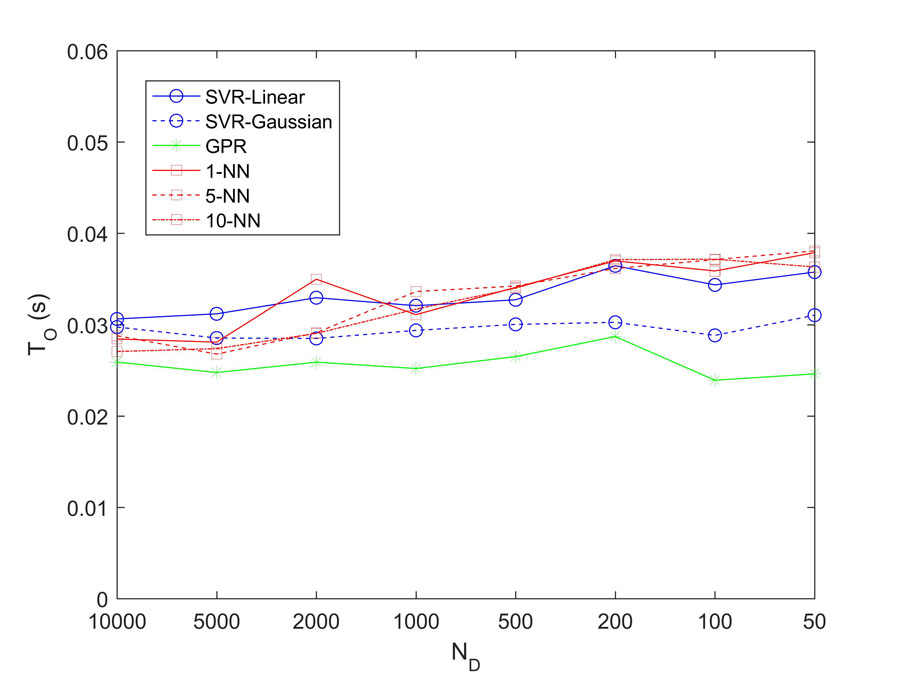}
\label{fig:rregression-tlo}
}
\caption{On-line learning time with different regression methods}
\label{fig:rregressiontl}
\end{figure}

To evaluate different regression methods, $k$-NN ($k=1, 5, 10$), SVR with linear kernel and Gaussian kernel and GPR are employed in the regression phase. We employ the second feature described in Section \ref{sec:case-feature}, refinement with objective function and database consisting $N_D$ samples $D_{N_D}$ drawing from the original database $D$, and build LO models $\mathcal{L}(D_{N_D},R_{k-NN},O)$, $\mathcal{L}(D_{N_D},R_{L-SVR},O)$, $\mathcal{L}(D_{N_D},R_{G-SVR},O)$ and $\mathcal{L}(D_{N_D},R_{GPR},O)$. The performance evaluated with $D_T$ are shown in Figure \ref{fig:rregression} and \ref{fig:rregressiontl}.

According to Figure \ref{fig:rregression}, for all the tested regression methods, little cost increase (less than $1\%$) and on-line motion time (less than 50ms) can be achieved, and high feasible success rate can be attained with large database size. Best feasibility and real-time performance can be achieved by GPR and followed by Gaussian-SVR. But the off-line model building time for GPR is larger than Gaussian-SVR, especially using database with larger size.

In general, feasible success rate decreases as the database size reduces. $k$-NN is most sensitive to the database size. When using fewer samples, the feasible success rate of $k$-NN decreases fastest, while GPR and Gaussian maintain a high feasibility, which is over $90\%$ with $N_D=50$. GPR has the best feasibility, which is over $99\%$ when $N_D>500$. Gaussian-SVR takes the second place, when $N_D>500$, $R_s>95\%$.

For GPR and SVR, regression models are built off-line, while only prediction and regression need to be performed in the on-line procedure, which takes little time for on-line calculation and leads to better real-time performance. The prediction times of GPR and SVR are much the same, but the regression time for GPR is less, then followed by Gaussian-SVR, and linear-SVR is the longest. In conclusion, GPR has the best real-time performance with the on-line calculating time less than 30ms. On the contrary, $k$-NN regression models are related to the input variables, thus required to be built on-line, which take longer time for on-line learning. In addition, the regression time becomes longer as the database size increases, for the reason that the similarity between the input and each sample need to be calculated. Learning with larger $k$ takes longer time in regression, but much fewer time in refinement, which makes the total learning time fewer.

For SVR, different kernel functions yield different results. Comparing with the linear kernel, Gaussian kernel yields higher feasible success rate, lower on-line learning time and more stable cost increase rate. Moreover, the time for model building of Gaussian-SVR is much fewer than linear-SVR.

For $k$-NN, larger $k$ yields higher feasible success rate, lower on-line learning time and stable cost increase rate.

In conclusion, very high feasibility and real-time performance can be attained by GPR with few samples. Otherwise, Gaussian-SVR is promising in model building time while can reach acceptable feasibility and real-time performance as well. $k$-NN has a conflict between feasibility and real-time performance.

\subsection{Parameter Refinement}
\begin{figure}[h]
\centering
\subfloat[On-line learning time]{\includegraphics[width=0.4\textwidth]{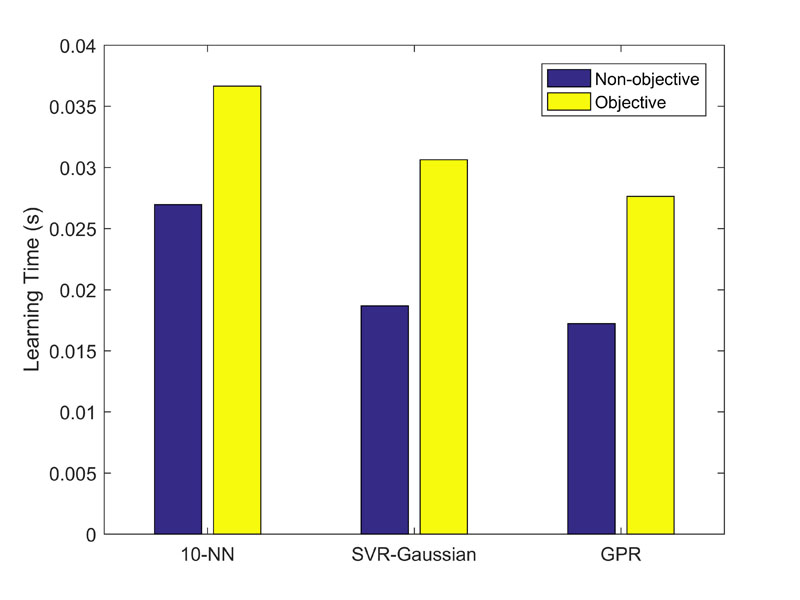}
\label{fig:rrefine-tl}
}
\hspace{1pt}
\subfloat[Errors]{\includegraphics[width=0.4\textwidth]{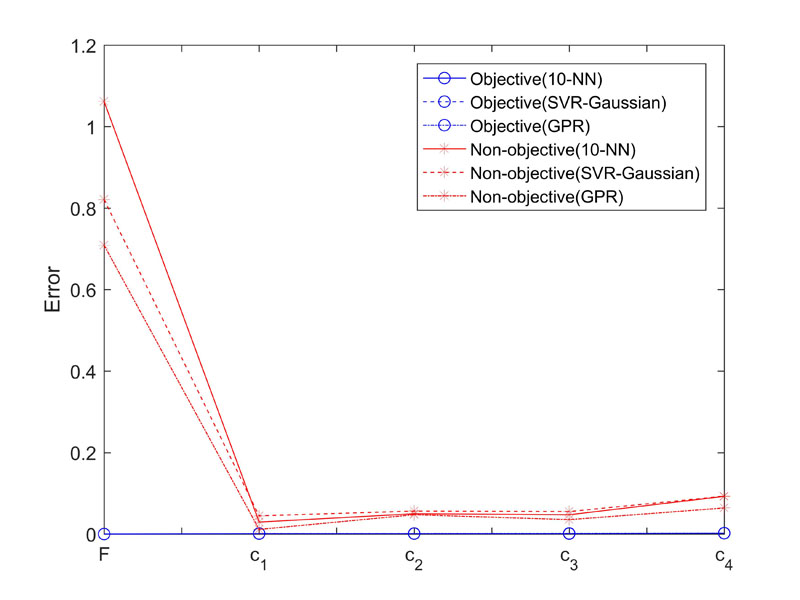}
\label{fig:rrefine-e}
}
\caption{Comparison of refinement with and without objective function}
\label{fig:rrefine}
\end{figure}

To evaluate the influence of the parameter refinement methods, refinement methods with and without objective function are employed, with database $D_{5000}$ and regression methods of 10-NN, Gaussian-SVR and GPR, to build LO models. The performances evaluated with $D_T$ are shown in Figure \ref{fig:rrefine}, which indicate that refinement with objective function costs fewer time in refinement stage but causes larger error of objective value.

\begin{figure}[h]
\centering
\subfloat[Cost increase]{\includegraphics[width=0.4\textwidth]{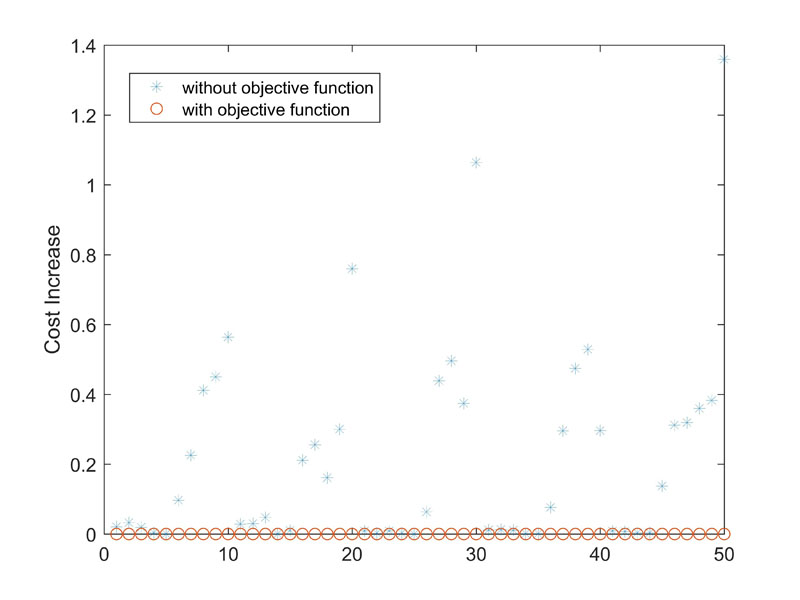}
\label{fig:rrefineerr}
}
\hspace{1pt}
\subfloat[Relationship of $e_F$ and ${dF}_r(\boldsymbol{C^\ast})$]{\includegraphics[width=0.4\textwidth]{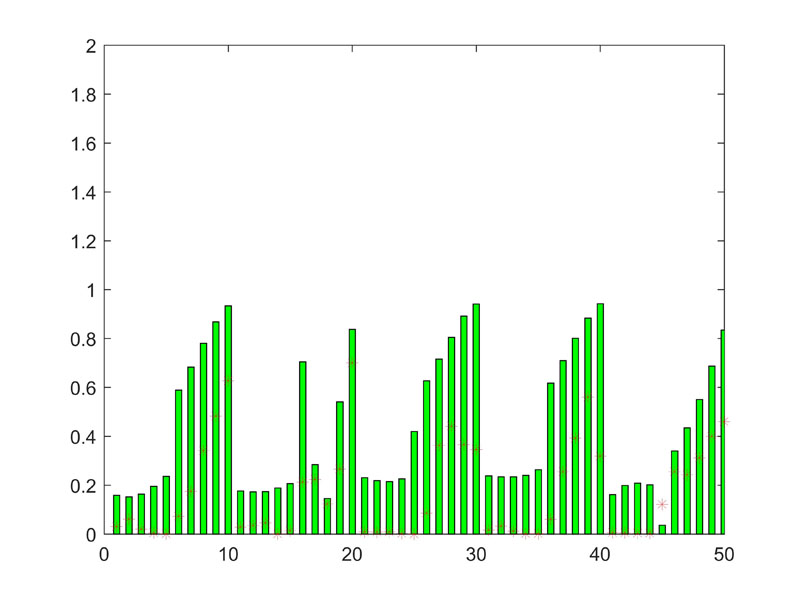}
\label{fig:rrefinedfr}
}
\caption{Cost increase of each datum}
\label{fig:rrefineerr}
\end{figure}

Examining cost increase rate of each datum (Figure \ref{fig:rrefineerr}), we find that the errors are tiny in most cases, but very large for several inputs, which lead to the high average error. The relative change rate at each $\boldsymbol{C^\ast}$ is calculated and plotted in Figure \ref{fig:rrefinedfr} as the green bars, which has a consistent trend of the error. That is, when ${dF}_r(\boldsymbol{C^\ast})$ is large, non-objective refinement tends to lead to a large cost increase rate though the parameter offset is very small.

To eliminate the cases with very large error, refinement with objective function is adopted, which significantly reduces the average error with a slight increase of refinement time. As the initial parameter obtained by regression is already close to the global minimum, the total learning time is still much smaller than the original optimization (105.3ms), and satisfies real-time requirement.

\subsection{Continuous Trajectory Planning}
\begin{figure}
\centering
\subfloat[Velocity]{\includegraphics[width=0.4\textwidth]{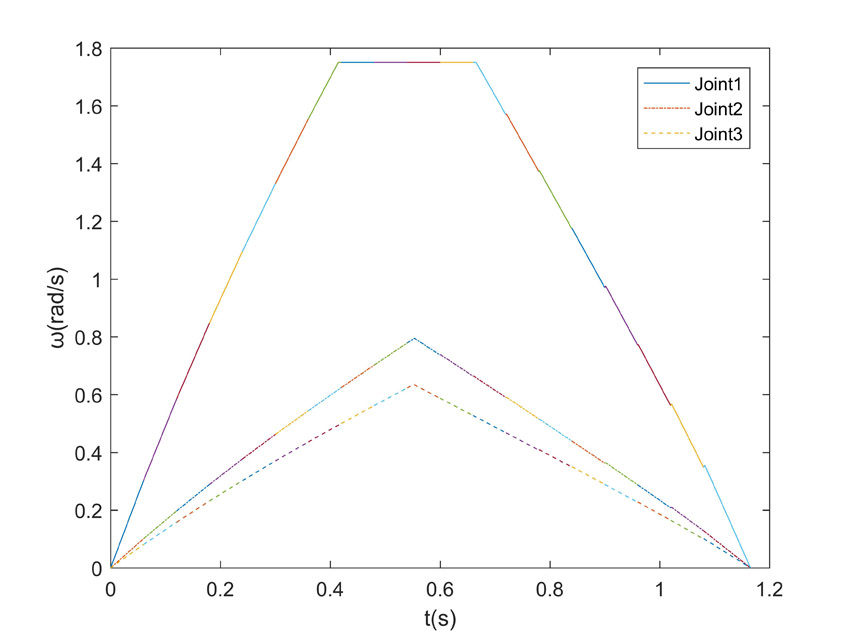}
\label{fig:rwholev}
}
\hspace{1pt}
\subfloat[Position]{\includegraphics[width=0.4\textwidth]{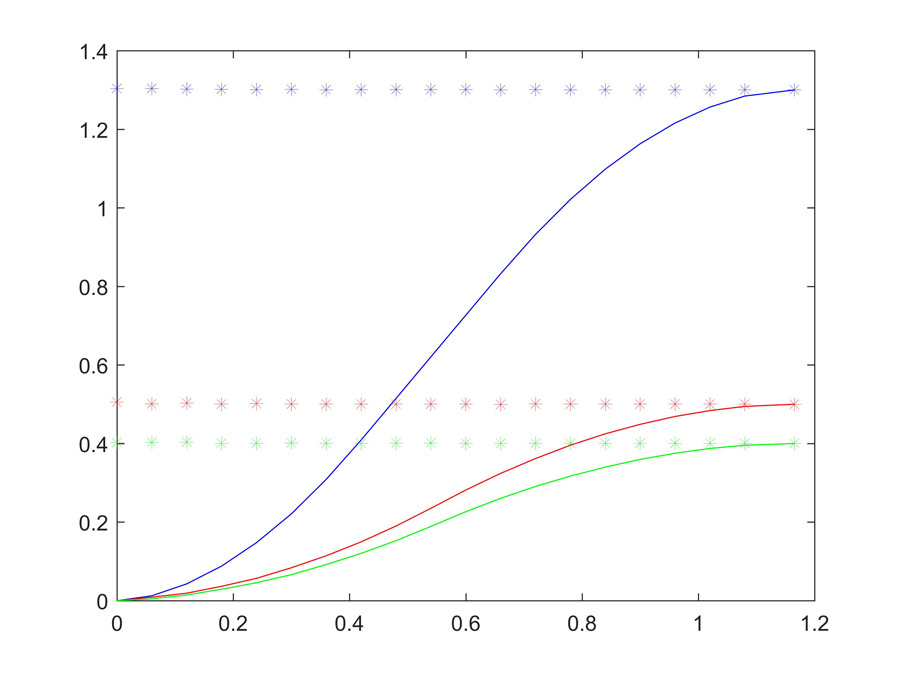}
\label{fig:rwholeq}
}
\caption{Motion curve during haptic feedback}
\label{fig:rrefineerr}
\end{figure}

The continuous trajectory planning for HFM is performed with the LO model $\mathcal{L}(D,R_{GPR},O)$ and the updating period $T_p=60ms$. The velocity curve of the whole process is shown as Figure \ref{fig:rwholev}. The joint offset curve and the predicted interacting position in each period are shown by the solid line and the stars in Figure \ref{fig:rwholeq}, respectively. Joint 1, 2 and 3 are represented by blue, red and green, respectively. By the LO method, smooth velocity curves can be obtained and each joint can finally reach the interacting point in the interacting time.

\section{Conclusions}
A real-time optimal trajectory generation method based on machine learning, together with the specific methodologies of off-line database generation and on-line learning for high dimensional inputs, are presented in this paper. Consequently, continuous trajectory planning of high-DOF-robot in complex environment is achieved. First, a learning optimization framework for continuous trajectory planning is constructed, in which different sub-methods include databases, regression methods and refinement methods are selected to adapt various requirements. In particular, aiming at high dimensional inputs, an input space dimension-reducing mapping method is presented. Database of high quality is obtained with higher feasible success rate of sample calculation and more reasonable sample distribution. And the database generation efficiency is elevated by reducing calculating time. Moreover, an on-line learning method based on multivariable-multiple-regression is presented. The learning efficiency can be improved by flexibly selecting features. Besides, Performance evaluation indices are defined, including feasible success rate, cost increase rate and learning time, which indicate the feasibility, real-time performance and accuracy of the LO model, respectively. Finally, in the case of haptic feedback, real-time continuous trajectory planning with high dimensional inputs is achieved by this LO method with database generating by the input space dimension-reducing mapping.

Further improvement of calculation efficiency of continuous trajectory planning with high dimensional inputs is needed, especially for situations with more trajectory parameters. Future work may focus on further improving calculation efficiency in both off-line and on-line procedure, for example, accelerating non-linear optimization by reducing times for off-line sample calculation and on-line regression.



\bibliographystyle{elsarticle-num}
\bibliography{continuous_trajectory_planning}


%
%
%

\end{document}